\documentclass[12pt]{article}
\usepackage{amssymb,amsmath}

\newcommand{\T}{\mathsf{T}}
\newcommand{\exd}{\mathrm{\mathbf{d}}}
\newcommand{\concept}[1]{``\textit{#1}''}
\newcommand{\rank}{\text{rank}\,}
\newcommand{\D}{\mathsf{D}}
\newcommand{\SE}{\mathsf{SE}}
\newcommand{\Sunit}{\mathsf{S}^1}
\newcommand{\Complex}{\mathbb{C}}

\usepackage{xr}
\usepackage[backend=bibtex,natbib=true,citestyle=alphabetic]{biblatex}
\usepackage[%hyperref
hypertexnames,%
citecolor=blue,%
colorlinks=true,%
linkcolor=red,%
bookmarks=true
]{hyperref}
\usepackage[all]{hypcap}
\usepackage{times}
\usepackage[section]{placeins}
\usepackage[pdftex]{graphicx}
\graphicspath{{figures/}}
\usepackage[bottom]{footmisc}
\usepackage{calc}
\usepackage[switch]{lineno}
\graphicspath{{figures/}}
\DeclareGraphicsExtensions{.pdf,.jpeg,.png}
\usepackage{steinmetz}

\usepackage{amsthm}
\usepackage{dsfont}
\usepackage{subcaption}
\usepackage{algorithm}
\usepackage[noend]{algpseudocode}
\usepackage{listings}
\usepackage{mathtools}
\theoremstyle{plain}
\newtheorem{theorem}{Theorem}
\newtheorem*{theorem*}{Theorem}

\theoremstyle{definition}
\newtheorem*{definition*}{Definition}
\newtheorem{definition}[theorem]{Definition}

\newcommand{\norm}[1]{\left\lVert#1\right\rVert}

\newcommand{\Reals}{\mathbb{R}}

\newcommand{\Image}[1]{\text{Im}(#1)}
\newcommand{\Pc}{\mathcal{P}}

\newcommand{\mref}[1]{M-\ref{#1}}

\usepackage{xcolor}
\usepackage{pdfpages}
\usepackage[margin=2cm]{geometry} 
\usepackage[backend=bibtex,natbib=true,citestyle=alphabetic]{biblatex}
\usepackage[
citecolor=blue,%
colorlinks=true,%
]{hyperref}
\usepackage[all]{hypcap}
\usepackage[all]{hypcap}
\usepackage{subcaption}
\newcommand\blfootnote[1]{%
	\begingroup
	\renewcommand\thefootnote{}\footnote{#1}%
	\addtocounter{footnote}{-1}%
	\endgroup
}

\begin{document}
\date{}
%\authormark{Cambridge Authors}

%\articletype{RESEARCH ARTICLE}

%\jnlPage{1}{00}
%\jyear{2021}
%\jdoi{10.1017/xxxxx}

%\author[1]{George Council\hyperlink{corr}{*}}
%\address[1]{Ford Motor Company, Dearborn, MI, USA}

%\author[2]{Shai Revzen}
%\address[2]{Department of Electrical Engineering, University of Michigan, Ann Arbor, MI, USA}
%\address{\hypertarget{corr}{*}Corresponding author. \email{gcouncil@umich.edu}}

%\keywords{robotics, recovery, constraints, templates, damage}
%\received{xx xxx xxx}
%\revised{xx xxx xxx}
%\accepted{xx xxx xxx}

\author{George Council \thanks{Ford Motor Company, Dearborn, MI, USA ({\tt gcouncil@umich.edu}). This work was completed under the support of the BIRDS Lab at the University of Michigan, and is in no way affiliated with Ford Motor Company. }
	\and
\and Shai~Revzen%
%\thanks{Department of Electrical Engineering and Computer Science, University of Michigan, Ann Arbor, MI, USA ({\tt shrevzen@eecs.umich.edu})}%
\thanks{EECS Dept., University of Michigan, Ann Arbor, MI, USA ({\tt shrevzen@eecs.umich.edu})}%
}
\title{Recovery of Behaviors Encoded via Bilateral Constraints}

\maketitle

\begin{abstract}
	If robots are ever to achieve autonomous motion comparable to that exhibited by animals, they must acquire the ability to quickly recover motor behaviors when damage, malfunction, or environmental conditions compromise their ability to move effectively. %
	We present an approach which allowed our robots and simulated robots to recover high-degree of freedom motor behaviors within a few dozen attempts. %
	Our approach employs a \concept{behavior specification} expressing the desired behaviors in terms as rank ordered differential constraints. %
	We show how factoring these constraints through an \concept{encoding templates} produces a recipe for generalizing a previously optimized behavior to new circumstances in a form amenable to rapid learning. %
	We further illustrate that adequate constraints are generically easy to determine in data-driven contexts.
	As illustration, we demonstrate our recovery approach on a physical 7 DOF hexapod robot, as well as a simulation of a 6 DOF 2D kinematic mechanism. %
	In both cases we recovered a behavior functionally indistinguishable from the previously optimized motion.
\end{abstract}

\tableofcontents
\blfootnote{ This work was supported by ARO
	grants W911NF-14-1-0573, W911NF-17-1-0243, and W911NF-17-1-0306, NSF CMMI 1825918, NSF CPS 2038432 and D. Dan and Betty Kahn Michigan-Israel Partnership for Research and Education Autonomous Systems Mega-Project. }
\section{Introduction}\label{sec:intro}
To associate the notion of \concept{autonomy} with a putative \concept{agent} implies that this agent has the ability to persist in carrying out its goals despite interference.
The more profound the changes in action it manifests to continue achieving its goal, the more autonomous we perceive that agent to be.

Most modern robots achieve autonomous motion by implementing a sense-plan-act loop in which they rely on accurate dynamic models, either derived from first principals or estimated from data, for predicting the consequences of potential actions.
When robots are damaged or the environment undergoes a large change, the accuracy of the predictions falters, and full replanning in real-time from scratch becomes infeasible because the models used for prediction cannot be reconstituted quickly enough.

Unlike robots, many animals display great aptitude at preserving motion despite changes in the underlying dynamics.
Through injury, age, or the otherwise mutable nature of organic tissue, animals are able to preserve behaviors despite great dynamic variability.
For example, in the case of light or moderate limb damage, we distinguish this resiliency from healing  --  a human with a sprained ankle immediately begins limping, rather than waiting for a healing process to restore the joint to full functionality.
Obviously, robots would be advantaged by a similar ability to preserve task execution through damage or changes in dynamics.

Since we wish to employ feedback control to compensate for this uncertainty, in contemporary geometric language, the model-based regime usually describe systems in terms of a differential generator -- the \concept{uncontrolled dynamics} -- and the space of possible control actions at each state -- the \concept{control distribution}.
The key insight behind our approach is the power of using the \emph{dual} to the conventional control theory approach.
We specify behaviors in terms of a (potentially over-determined) ranked list of differential constraints.

When considering the spectrum of dynamical systems subject to constraints, a class of robots that has received considerable are those than can be expressed as a collection of bilateral constraints -- most conspicuously, mechanical systems expressed as collections of rigid bodies subject to kinematic constraints  \cite{murray2017robotmanipulation}; equivalently, masses or links connected by revolute and prismatic joints that may or not be actuated, and similarly, though not exhaustively, such mechanisms subject to grasping constraints \cite{murray2017robotmanipulation}, nonholonomic constraints \cite{bloch2003nonholonomic} or soft contacts \cite{eldering2016role}.
Very often, the constraints are resolved into a differential equation (the aforementioned differential generator) arising from the Euler-Lagrange equations subject to the correct constraint regime.

As such, many types of mechanical failure correspond to the removal or addition of kinematic constraints: e.g. a wheel losing traction removes a constraint, or a bearing locks, adding a new constraint.
We will show how in such cases our ranked list of constraints can be used to naturally define a control problem with a richly informative local gradient.
We will then also show that this approach is amenable to both explicit constructive solutions when underlying dynamic models are available, as well as defining an optimization problem that be solved via shooting approaches in hardware.
%The learning problem, enhanced by inclusion of additional ``natural'' constraints, converges much more rapidly than methods that lack this rich local information.

In our representation, high-ranking differential constraints represent unbreakable physical constraints imposed by physics and mechanical structure, whereas the lower-ranking constraints represent design choices and priorities.
Whether constraints appear, disappear, or change, we always employ the dominant list of constraints to choose the robot's action, thereby utilizing as much state-local information as possible.
This mathematical representation makes it particularly easy to specify a desired behaviors using the conceptual framework of \concept{Templates and Anchors} \cite{full1999templates}, \cite[Chp.3]{sharbafi2017bioinspired}.

The template and anchor framework claims that in biology the movements of high degree of freedom \concept{anchor} models that closely reflect the structure of the animal body additionally follow motions of a lower dimensional \concept{template} model in which anchor degrees of freedom are coupled together tightly.
We employ this insight by first selecting a lower dimensional collection of outputs which can reliably describe the desired outcomes, which we named the \concept{encoding template}.
We then specify the behavior in terms of differential constraints on the encoding template, and pull those constraints back to the anchor system, thereby defining a  non-holonomic system on what is, typically, the physical configuration space of the robot.
The utility of this construction is that \emph{any} anchor trajectories which satisfy these constraints map to the desired template behavior.
For example, ``walk straight across a room'' could be expressed in terms of projection of the center-of-mass (CoM) on the horizontal, followed by constraining the CoM motion to be on a family of parallel lines, and also to have chosen a non-zero CoM velocity in the desired direction.
Such a constraint on the projection is not a complete definition for a gait of a non-trivial legged robot, but any motion that met these differential constraints would accomplish the objective of moving across the room.

The power of our dual constraint-based representation is that adding these designed constraints to the existing immutable physical constraints of the robot consists of merely concatenating the designed constraints on the end of the list of physical constraints.
Additionally, the constraints which define the chosen encoding template arise as \emph{outputs} -- their image value dictates whether or not the constraint is satisfied, naturally defining a \emph{prior}.
Ergo, encoding constraints can be used \emph{descriptively} on an example execution to decide if that execution satisfied a desired constraint or not.
Furthermore, when an ensemble of desirable anchor trajectories is available for training, a collection of data-driven encoding template constraints can easily be learned, added as low priority constraints, and used to assist the recovery of similar behaviors when needed.

There have been numerous authors that have considered the control of constrained dynamical systems with such geometric methods.
Principal fiber bundles \cite{bloch1996nonholonomic},
Kinematic reduction \cite{bullo2002controllable}, spanning killing forms \cite{bloch1995another}, and other techniques from Riemannian geometry \cite{vershik1972differential, synge1928geodesics, lewis1998affine} (and the references therein), to name a few, are all powerful and general approaches to non-holonomic motion planning.
Our approach uses less structure, and while our mathematical results are correspondingly weaker, we have an advantage in practice -- the ability to learn constraints which produce a specific desirable trajectory, using a nearly arbitrary choice of outputs taking values in the encoding template.
Thus our approach requires little to no model information to recover template trajectories, whereas the previous techniques typically rely on the availability of a precise model.
In the work presented here we further differentiate ourselves from other constraint-based work in that we only attempt to recover a training example, although we could in principle extend to more general classes of training data.

When it comes to instantiation of reduced-dimension templates, previous work, e.g. especially of inverted pendulum reductions in bipedal walkers \cite{xiong2018coupling, griffin2017nonholonomic, poulakakis2007formal} has delved into the much more challenging problem of planning on a template, which requires that open neighborhood of  template trajectories must be liftable to the anchor.
Because we only required that a single chosen trajectory be lifted, this greatly reduced the amount of structure we needed to impose on the problem.
Below we demonstrate that our gait recovery technique is computable using data-driven methods applied to physical robots, and works rapidly in practice.

\subsection{Comparison to Other Work}
Reinforcement-learning based approaches for  damage-compensation on legged robots has been explored previously, notably in \cite{cully2015robots} with intelligent trial-and-error, and \cite{bongard2011morphological} with continuous self-modeling. 
We distinguish our approach as we do not require (or fit) a predictive model,  dynamic or otherwise (such as a neural network),  at any stage, nor do we need to perform extensive computation -- our experiments indicate that a local gradient descent on input, i.e., for a parameterized class of functions $u(\alpha)$ that drive the robot yields adequate local performance without global or semi-global performance knowledge.
Owing to the \emph{assumption} that a failure results in a low-rank change in constraints renders the intrinsic dimension of recovery small, and likely localized (given adequate actuator freedom), even though a predictive model could have drastically different coefficient functions if a constraint is changed (e.g., consider how the Euler-Lagrange equations would change as functions for the addition or removal of constraints when expressed in minimal coordiantes).
We also remark that our notion of ``fast'' is in \emph{real time} -- approaches such as PILCO \cite{deisenroth2011pilco} that demonstrate small \emph{interaction time} can have an burdensome offline computation that renders them slow in practice, even though they make extremely efficient use of data. 
Similarly, reinforcement learning approaches for multi-task robots with configurable geometry, e.g. (though hardly exhaustive), via graph neural networks \cite{wang2018nervenet,whitman2021learning},  prior experience \cite{yang2020data}, and grammars \cite{zhao2020robogrammar} are subject to similar computation requirements in an expensive training stage, obstructing real-time performance. 

Our primary contention is that while these approaches certainly demonstrate remarkable ability to elicit effective motion from variable robot geometries, they consume significant computational resources, and lack \emph{explainability}.
Principals or structure that might underlay the recovery problem are obscured by the black-box nature of neural networks or other dense functional representations.
The constraint-based formulation we present in the sequel, when formulated as a optimization problem via. \eqref{eqn:con-cost-function} seems to demonstrate double-digit iteration count, relies on zero-order optimization methods, yet recovers motion in 30 mins of real time. 
The performance we observe may be due to the straightforward nature of the general structure we identify, and furthermore, that structure may ultimately be helpful in determining or explaining the high-performance of machine learning methods that otherwise lack formal guarantees. 

Furthermore, our constraint-based observation need not exist in opposition from reinforcement learning methods.
The constraints define a natural cost (\eqref{eqn:con-cost-function}) that can be used as a reward function \cite{achiam2017constrained}, but it is more compelling to consider appending learned constraints to an a priori putative cost function employed to re-learn motion on a damaged robot.
While we do not test that particular formulation in this manuscript, there is some empirical evidence to suggest that reinforcement learning methods for path-planning  can have greatly improved convergence when subjected to constraints in this manner \cite{nageshrao2019autonomous,ericrobustdriving}, or that ``heuristics'' (which can be interpreted as constraints) can be used to summarize complex dynamic behavior of legged robots to improve optimization speed \cite{bledt2020extracting}.

\subsection{Mathematical preliminaries}\label{sec:math-background}

We assume that a robot's motion is determined by curves $x(t)$ taking values in a manifold $Q_R$ on which we can write differential constraints.
Typical choices for $Q_R$ could be the configuration space of the robot body, its phase space \cite{arnold2013mathematical}, or a more general state space.
While it may seem initially strange to allow the domain to vary so generally, we observe that a tangent vector \cite{lee2013smooth}, which is our fundamental object of interest, is naturally defined in an equally general setting.

We define a \concept{behavior specification} to be a list of constraints of the form $\Omega_i (x) \cdot \dot{x}=\gamma_i(t,x)$, $i=1,..., k$.
Here each $\Omega_i$ is a differential 1-form \cite{lee2013smooth} i.e. a section of the cotangent bundle $\T^*Q_R$.
The vector $\gamma(t,x):=(\gamma_1, \dots, \gamma_k)$ is a vector of length $k$ that defines the value the constraint functions $\Omega_i$ must satisfy, and therefore takes values in the same codomain as that of the 1-forms.
The list of $\Omega_i$ contains any inviolate physical constraints that determine the physics of the robot, as well as constraints we as designers wish to engineer into the system.
Conventionally, the constraints could be written as a matrix: $\Omega(x)\cdot \dot{x} = \gamma(t,x)$, where the rows of $\Omega$ are the $\Omega_i(x)$.
We assume the matrix $\Omega$ to be of constant, though necessarily not full, rank when evaluated along admissible curves of $Q_R$.
Formally speaking, the behavioral specification is the pair $(\Omega,\gamma)$.

A curve $x(t)$ taking values in $Q_R$ which satisfies the behavior specification equation is an instance of the behavior.
A major feature of our behavior representation is that it is agnostic of the mechanism that generates curves $x(t)$.
In particular, instances of the behavior may intersect and even overlap, only to diverge later -- unlike trajectories of conventional closed loop control models.
With respect to a behavior specification there are only curves that satisfy the constraints, and those that do not.

We chose this definition for a behavior as it contains a number of special cases.
For example, if $\Omega$ is invertible everywhere, we may write $\dot x = \Omega^{-1}(x) \gamma(t, x)$ -- a conventional non-autonomous ordinary differential equation (ODE).
Here instances of the behavior are solutions of the ODE.
The popularly used class of affine control systems $\dot{x}= f(x) + G(x) u$ can be represented using a pseudoinverse $G^\dagger$ of $G$, by constructing $\Omega(x) := \mathrm{I} - G(x) G^\dagger(x)$ and $\gamma(t,x) := \Omega(x) f(x)$, 
i.e., those components of $\dot{x}$ which cannot be directly controlled by the input $u$ must agree with the drift-term $f$ projected into the corresponding directions.
This is a standard application of control redesign.
If the space $Q_R$ is taken as the configuration space of a mechanical system, Pfaffian and affine differential constraints are behavior specifications as well.
In kinematic reduction, the constraints would be the metric inner products of \cite{bullo2002controllable}.
The preceding list of model types that can be realized as behavior specifications is not exhaustive, but is intended to indicate that a number of useful constructions of control and robotics are naturally encapsulated in our proposed definition.

While the constraints required by physics are intrinsic to the system, it is not immediately clear from the definition given how to encode a design goal into a collection of constraints.
We propose the following strategy: we first find a manifold $Q_E$ and a full-rank function $\phi:Q_R \to Q_E$ such that we are certain that whatever outcome we desire is realized by a behavior specification on $Q_E$.
We call the space $Q_E$ an \concept{encoding template}, as it encodes all the necessary information.
Generally, we take the dimension of $Q_E$ to be less than that of $Q_R$.
The map $\phi$ reduces the coordinates of $Q_R$ to values we as designers care about encoding; for example, $\phi$ could return the CoM coordinates, an end effector location, joints angles, etc.
The map $\phi$ can equally be considered a collection of \concept{outputs} $y_i := \phi_i(x)$.
We write the behavior specification $\omega_i(y) \cdot \dot{y}=\eta_i(t,y)$ on the output variables / encoding template.
Such a construction precisely includes all the special cases a behavioral specification can capture, but only on the output variables.
We can then pull the $\omega_i$ back to $Q_R$ to augment any extant $\Omega$ by $\Omega_j$.
In matrix form, pulling back is merely adding rows $\Omega_j(x) := \omega_i(\phi(x)) \cdot \mathsf{D}\phi(x)$ to the matrix $\Omega$.
We augment our notation to include $\phi$ in the behavior specification as the tuple $(\phi, \omega, \eta)$, indicating that it is the image of $\phi$ which is the target of our design efforts, emphasizing that there are virtual constraints on the output variables in concert with pre-existing constraints that are defined only on $Q_R$.
The requirements on the map $\phi$ are substantially relaxed from the \emph{asymptotic} phase requirements of the templates described in \cite{full1999templates}, which are better known. 
For a expanded discussion of how an encoding template is distinct from an aysmptotic template, refer to appendix A. 

\subsection{Recovery via constraints} \label{sec:recovery-constraints}
From here on we restrict our interest to the recovery of a behavior on a robot post-disruption.
We assume that there was a distinguished curve $x_0(t) \subset Q_R$ that satisfied a given behavior specification $(\phi, \omega, \eta)$.
For emphasis, we will consider the case where the number of constraints \emph{exceeds} the dimension of $Q_R$, but that the virtual constraints $\omega$ are satisfied without control effort.
In this case, the rows defined by $\omega$ are redundant with the rows of $\Omega$ -- the rank of $\omega$ augmenting $\Omega$ is identical to that of $\Omega$ alone along desired trajectories in $Q_R$.
Thus, at this point we have three classes of constraints: $\Omega_P$ that come from the underlying physics, $\Omega_D$ which are design constraints derived from the $(\phi,\omega,\eta)$ specification, and $\Omega_L$ constraints that were learned from the encoding of the example $x_0(t)$, i.e. by observing $\phi(x_0(t))$.
These constraints can be viewed as if they are enforced in an order of priority $\Omega_P \gg \Omega_D \gg \Omega_L$; here we indicate priority by $\gg$.

We assume that the robot is disrupted in a manner that introduces a new $\Omega_r$ to $\Omega_P$, or eliminates one of the native $\Omega_i$ that comprise $\Omega_P$, representing effects such as motors seizing, limbs breaking off, etc.
The recovery strategy is to re-enforce, via control, the design constraints $\Omega_D$, which are presumably violated by whatever motion the broken robot is performing without compensation.
If the rank of $\Omega_P$ was reduced, the learned constraints $\Omega_L$ which were originally redundant, can play an essential role in completing the behavior specification to full rank.

We have assumed $\dim Q_E < \dim Q_R$, and that the behavior specification $(\phi,\omega,\eta)$ is satisfied by the example trajectory $x_0$, which was presumably obtained from a computationally intensive offline optimization.
Thus we know:
\begin{equation}\label{eqn:x0}
\forall j,t:~ \omega_j(\phi(x_0(t))) \cdot \D \phi(x_0(t))\cdot \dot x_0(t)
= \eta_j(t, \phi(x_0(t)))
\end{equation}
From the constant rank assumption about $\phi(\cdot)$ we obtain that for each $x_0(t)$ there is an entire manifold of possible values for a new instantiation $x(t)$ such that $\phi(x(t))=\phi(x_0(t))$.

Before damage, we took $(\Omega,\gamma)$ as $(\Omega_P,\gamma_P) \gg (\Omega_D,\gamma_D) \gg (\Omega_L,\gamma_L)$ and used the first $\dim Q_R$ linearly independent constraints of these determine the velocity $\dot x(t)$.
However, by virtue of the addition of $\Omega_L$, the total number of constraints in $(\Omega,\gamma)$ is larger than $\dim Q_R$ (i.e. $\Omega$ is tall), and these constraints are redundant on the instantiation of the behavior $x_0(\cdot)$.
As long as the robot was functioning without damage, the $\Omega_D(x_0) \cdot \dot x_0 = \gamma_D(t,x_0)$ constraints were satisfied by assumption, and no change in control was needed.

Damage to the robot was a low rank change to $\Omega_P$, replacing it with $\tilde{\Omega}_P$ instead, of possibly lower or higher rank, but such that only a few constraints are affected.
In other words, only a few rows of $\Omega_P$ and $\gamma_P$ are changed due to damage.
Consider the case where the rank change, i.e. change in number of constraints, associated with this damage to $\Omega_P$ is such that
\begin{equation}\label{eqn:dmgRnk}
\dim Q_R - \rank \Omega_L \leq \rank \tilde{\Omega}_P + \rank \Omega_D \leq \dim Q_R
\end{equation}
When \eqref{eqn:dmgRnk} holds, the change in rank induced by the damage can be taken up by removal or addition of learned constraints $\Omega_L$, and we could solve for new feasible velocities without harming compliance with any of the design constraints $\Omega_D(x) \cdot \dot x = \gamma_D(t,x)$.
\begin{figure}[ht]
	\centering
	%% Creator: Inkscape 1.0.2 (e86c870879, 2021-01-15), www.inkscape.org
%% PDF/EPS/PS + LaTeX output extension by Johan Engelen, 2010
%% Accompanies image file 'connection.pdf' (pdf, eps, ps)
%%
%% To include the image in your LaTeX document, write
%%   \input{<filename>.pdf_tex}
%%  instead of
%%   \includegraphics{<filename>.pdf}
%% To scale the image, write
%%   \def\svgwidth{<desired width>}
%%   \input{<filename>.pdf_tex}
%%  instead of
%%   \includegraphics[width=<desired width>]{<filename>.pdf}
%%
%% Images with a different path to the parent latex file can
%% be accessed with the `import' package (which may need to be
%% installed) using
%%   \usepackage{import}
%% in the preamble, and then including the image with
%%   \import{<path to file>}{<filename>.pdf_tex}
%% Alternatively, one can specify
%%   \graphicspath{{<path to file>/}}
%% 
%% For more information, please see info/svg-inkscape on CTAN:
%%   http://tug.ctan.org/tex-archive/info/svg-inkscape
%%
\begingroup%
  \makeatletter%
  \providecommand\color[2][]{%
    \errmessage{(Inkscape) Color is used for the text in Inkscape, but the package 'color.sty' is not loaded}%
    \renewcommand\color[2][]{}%
  }%
  \providecommand\transparent[1]{%
    \errmessage{(Inkscape) Transparency is used (non-zero) for the text in Inkscape, but the package 'transparent.sty' is not loaded}%
    \renewcommand\transparent[1]{}%
  }%
  \providecommand\rotatebox[2]{#2}%
  \newcommand*\fsize{\dimexpr\f@size pt\relax}%
  \newcommand*\lineheight[1]{\fontsize{\fsize}{#1\fsize}\selectfont}%
  \ifx\svgwidth\undefined%
    \setlength{\unitlength}{277.79527559bp}%
    \ifx\svgscale\undefined%
      \relax%
    \else%
      \setlength{\unitlength}{\unitlength * \real{\svgscale}}%
    \fi%
  \else%
    \setlength{\unitlength}{\svgwidth}%
  \fi%
  \global\let\svgwidth\undefined%
  \global\let\svgscale\undefined%
  \makeatother%
  \begin{picture}(1,1.30612245)%
    \lineheight{1}%
    \setlength\tabcolsep{0pt}%
    \put(0,0){\includegraphics[width=\unitlength,page=1]{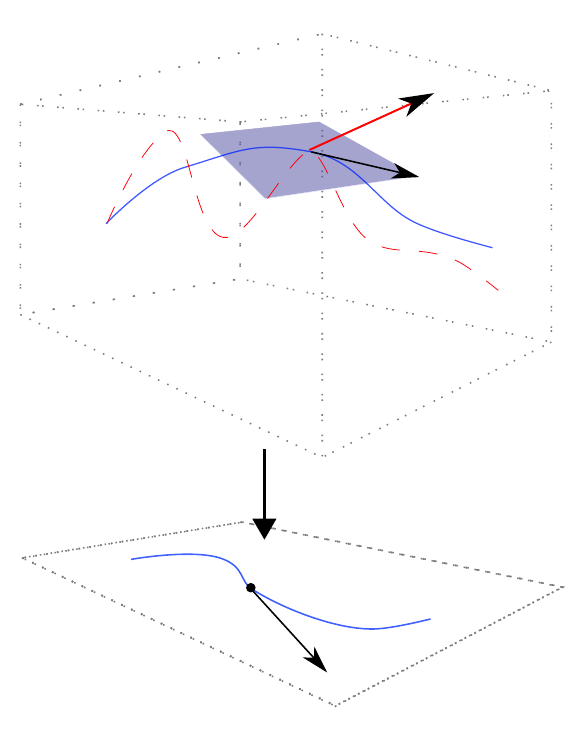}}%
    \put(0.40346006,0.44716612){\color[rgb]{0,0,0}\makebox(0,0)[lt]{\lineheight{1.25}\smash{\begin{tabular}[t]{l}$\varphi$\end{tabular}}}}%
    \put(0.49835437,0.27819226){\color[rgb]{0,0,0}\makebox(0,0)[lt]{\lineheight{1.25}\smash{\begin{tabular}[t]{l}$y(t)$\end{tabular}}}}%
    \put(0,0){\includegraphics[width=\unitlength,page=2]{connection.pdf}}%
    \put(0.24407823,0.76101114){\color[rgb]{0,0,0}\makebox(0,0)[lt]{\lineheight{1.25}\smash{\begin{tabular}[t]{l}$\ker \Omega$\end{tabular}}}}%
    \put(0.89764429,0.86800805){\color[rgb]{0,0,0}\makebox(0,0)[lt]{\lineheight{1.25}\smash{\begin{tabular}[t]{l}$x(t)$\end{tabular}}}}%
    \put(0.87664448,0.74964578){\color[rgb]{0,0,0}\makebox(0,0)[lt]{\lineheight{1.25}\smash{\begin{tabular}[t]{l}$\tilde{x}(t)$\end{tabular}}}}%
    \put(0,0){\includegraphics[width=\unitlength,page=3]{connection.pdf}}%
    \put(0.58181329,0.38953026){\color[rgb]{0,0,0}\makebox(0,0)[lt]{\lineheight{1.25}\smash{\begin{tabular}[t]{l}$ D \varphi^{\dagger}$\end{tabular}}}}%
  \end{picture}%
\endgroup%

	\caption{A schematic representation. The anchor (upstairs) and the template (downstairs) are related by projection $\varphi$.  The differential section $D\varphi^{\dagger}$ must be chosen to lie in the affine subspace $\Omega^{-1}{\gamma}$, which is free  in $\text{ker} \Omega$. Curves $x(t)$ and $\tilde{x}(t)$ that satisfy the constraints both project to $y(t)$.}
\end{figure}
Here the use of the dual representation in our behavior specification came into its own.
It allowed us to gracefully recover from structural changes in the constraints governing the robot.
If an explicit form of $\tilde{\Omega}_P$ is known, finding a recovery trajectory requires no optimization to be done -- it follows directly from integrating the new behavior specification equation with the modified constraints; we demonstrated this in \S \ref{sec:crawler} below.

In the case of a physical robot, the modified constraints might not be known; we explored this possibility by attempting to re-learn a walking behavior for a hexapedal robot.
For this, optimization is a natural tool.
Taking a motion over time $t \in [0,T]$, using control input $u(t)$, and producing trajectory $x(t)$, it is common to express cost as an integral.
For a behavior specification, this suggests a natural choice of cost function:
\begin{equation}\label{eqn:con-cost-function}
J[x,u] := R[u]+  \lambda\int_0^T \|\Omega_{D,L} (x) \cdot \dot{x} -\gamma_{D,L}(t,x)\|^2\,dt
\end{equation}
The function $R$ penalizes the input, while the remainder penalizes the failure to meet the behavior specification (in the $\Omega_D,\gamma_D$ part) and penalizes any other discrepancies from the encoding of the example (in the $\Omega_L,\gamma_L$ part).
The constraints explicitly measure and penalize directions of $\dot{x}$ which, by virtue of carrying through $\phi$ are deemed relevant.

We used this approach in our hardware-in-the-loop optimization in section \ref{sec:Enepod}.
In this task we were only concerned with the end-point of the robot motion and therefore we took the control cost functional $R=0$.
For a manually tuned gait which walked forward, we learned a behavior specification $(\Omega_L, \gamma_L)$ using a choice of encoding template motivated by the Lateral Leg Spring (LLS) \cite{schmitt2000mechanical} and Spring Loaded Inverted Pendulum (SLIP) \cite{blickhan1989} dynamic templates.
We then initialized the optimization with a gait that left the robot stationary, and ran it with the violation of constraints norm and no end-point goal.
This optimization allowed our robot to re-learn an effective forward gait in 36 iterations.

\subsection{Relation to geometric mechanics}

An important special case which motivated much of our work was the case where the output variables can be split into components $y=(s,g)$, with the $s$ variables being the directly controllable robot \concept{shape variables}, and the $g$ variables being controlled through the intermediate action of $s$, $\dot{s}$, and the constraints.
Typically, $g$ represents global position and orientation of the robot or of an object the robot is manipulating.

The behavior specification constraints can be written as
\begin{equation}\label{eqn:con-def}
\omega(s,g) \cdot (\dot{s}, \dot{g}) = \gamma(s,g)
\end{equation}
When the space $Q_E$ is a configuration space and $\gamma(s,g) = 0$, equation \eqref{eqn:con-def} coincides with the familiar Pfaffian-constraint case.

The application of the forms $\omega(s,g)$ to $(\dot{s}, \dot{g})$ can, from linearity, always be re-written as $\omega_s(s,g)\cdot \dot{s}+ \omega_g(s,g)\cdot \dot{g}$, where $\omega_s$ and $\omega_g$ are left and right blocks of the matrix form of $\omega$ with $\dim s$ and $\dim g$ columns respectively.

If $\omega_g$ has a left inverse $\omega_g^\dagger$, we can obtain for any given $s(t)$:
\begin{equation}\label{eqn:reconstruction}
\dot{g} = \omega_g^\dagger(g,s) \left(\gamma(t,g,s) - \omega_s(g,s) \cdot \dot{s} \right),
\end{equation}
a non-autonomous differential equation for $g$.
Thus, enforcing the constraints of equation \eqref{eqn:con-def} uniquely determines the curve $g(t)$ from $s(t)$ and the initial $g(0)$.
In other words, if we maintain the related $(\Omega_D,\gamma_D)$, it ensures that the robot moves the same way through space, or manipulates the object it is moving in the same way.
The typical case where such a \concept{reconstruction equation} appears is in the \concept{non-holonomic connection} -- see Appendix D for more details, as well as  \cite{ostrowski1998geometric, bloch2003nonholonomic, bloch1996nonholonomic, koon1997geometric, hatton2011introduction}, and the references therein.

\section{Results}

\subsection{Manipulator Equation}\label{sec:EL-lambda}

To ground the preceding dialogue, we consider a straightforward special case from mechanics -- eliding the encoding template or learned constraints, we highlight the particular features of constraints we argue to exploit.
Smooth mechanical systems subject to bilateral Pfaffian holonomic or non-holonomic, via the Lagrange-d'Alembert principle \cite{bloch2003nonholonomic}, have dynamics that can expressed as a second-order differential equation with Lagrange multipliers. 
More precisely, for configuration $q \in \mathbb{R}^n$, constriants $A(q) \cdot  \dot{q}= 0$, control input $u \in \mathbb{R}^r$, and $\lambda \in \mathbb{R}^m$, we may write:
\begin{equation}\label{eqn:EL-cons}
M(q) \ddot{q} + C(q,\dot{q}) = B u + A(q)^T \lambda
\end{equation}
Where $\lambda$  are the coefficients of the constraint forces, $M(q) \in \mathbb{R}^{n \times n}$ is the inertia tensor matrix, $C(q,\dot{q}) \in \mathbb{R}^n$ is the aggregate of Coriolis and potential terms, and $B \in \mathbb{R}^{n \times r}$ is the control input mapping.
For full details, especially of the functional relationship between $\lambda$ and $q$, see a standard text on mechanics, e.g, \cite{murrayBook} or \cite{bloch2003nonholonomic}.

Suppose then that we posses a distinguished trajectory $x_d(t), ~t \in [0,T]$ with associated control input $u_d(t), ~t \in [0,T]$. 
Associated to this pair are the resulting constraint forces $\lambda_d(t)$ that result from solving \eqref{eqn:EL-cons} subject to the constraints. 
Defining $\eta(u,\lambda) := Bu + A(q)^T \lambda$, we obtain an associated ``collection of forces'' $\eta_d(t):=\eta(u_d(t), \lambda_d(t))$  along the trajectory $x_d$. 

Consider now that the constraints $A(q) \cdot \dot{q}$ where modified so that $A \mapsto A + \Delta$ where $\Delta$ is a perturbation that is rank-preserving.
\footnote{Here we imagine $\Delta$ being composed of both a rank-destroying alteration (damage) combined with a desired ($\Omega_D)$ or learned ($\Omega_L$) constraint to restore the desired rank.}
If we can solve the control redesign problem so that $Bu(x,t) + (A+\Delta)^T \lambda = \eta_d(t)$ for a trajectory $x_1(t)$ emanating from the initial condition $x_d$, then via the uniqueness of solutions for smooth vector fields, it must be that $x_1(t) = x_d(t), ~t \in [0,T]$ as, simply, the R.H.S of \eqref{eqn:EL-cons} is the same between the original and perturbed systems.

Two observations are particularly germane.
The first is that only the signal $\eta(u,\lambda)$ is required for reconstruction, and it does not depend on the model information of $M$ and $C$.
The second is that we do not need to know $A$ or $\lambda$  as given by \eqref{eqn:EL-cons} -- physically, we do need to restrict ourselves to knowing constraints in a preferred set of units.
E.g., if we are only able to measure linear combinations of constraint forces $Q \cdot A^T$ for an invertible matrix $Q$, and we instead defined $\xi = QBu(x,t) + Q A^T \lambda$ the recovery process would still be to redesign $u$ so that $\xi$ was preserved, as it uniquely determines $\lambda$. 
Indeed, we should not find this surprising, as the Euler-Lagrange equations are naturally coordinate-invariant, even though the resulting units may lack physical significance.

\subsection{Crawler}\label{sec:crawler}

\begin{figure}[ht]
	\begin{subfigure}{0.45\textwidth}
	\centering
	\includegraphics[width=0.8\linewidth]{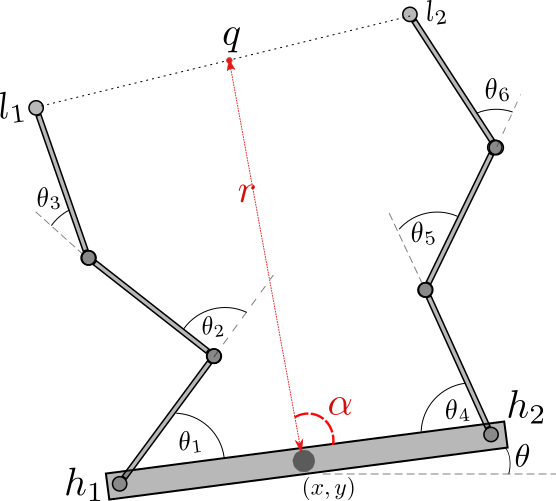}
	\caption{}
	%\caption{}
	\label{fig:crawler-sch}
    \end{subfigure}
	\begin{subfigure}{0.45\textwidth}
	\centering
	\includegraphics[width=\linewidth]{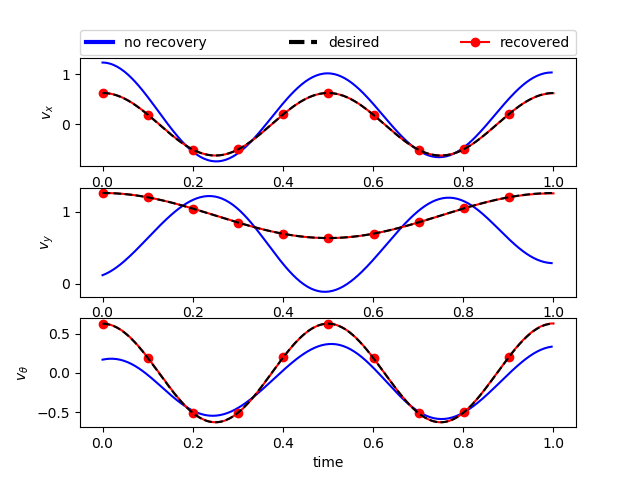}%{crawler_tracking_3.png}
	\caption{}\label{fig:crawler-tracking}
	\end{subfigure}
	\caption{ %
		(a)
		Illustration of our crawler. %
		The grey members indicate components that belong to the robot. %
		The points $l_1$ and $l_2$ are fixed foot locations. %
		Each joint $\theta_j$ is a powered rotational joint. %
		The points $h_1$ and $h_2$  are the attachment points at which the limbs attach to the body. %
		The center point of the foot locations $q$ defines the value $r$ and angle $\alpha$ (in red), which are our choice of encoding template.
		(b)
		The COM group velocity $\dot{g} = (v_x, v_y, v_{\theta})$ over time. %
		The ``desired'' curve (black dashed) is the undamaged motion that we are aiming to recover. %
		The ``old'' curve (blue solid) is the group velocity achieved if no recover strategy is attempted post damage. %
		The ``recovered'' trace (dots; red solid) is the performance after recovery.
    }
\end{figure}

To illustrate the step-by-step nature of our procedure, we present a simulated two-armed robot pulling itself on a plane. 
Since many of the resulting computations will be familiar to the reader, see the appendix for fine details.
In this section, we assume that all constraints and model information is known. 
In such a case, a six-dimensional non-autonomous ordinarily differential equation can be constructed -- its solution curve from the given initial condition yields the required joint trajectories for the damaged robot to preserve its encoding template trajectory. 
The robot is depicted in Fig. \ref{fig:crawler-sch}.
Each arm is a linkage that consists of four rigid bars connected end-to-end by powered swivel joints.
Our objective was to preserve the motion of the body when one of the joint actuators is jammed.
We implemented this example in Python 2.7.5 with the \verb|numpy| and
\verb|scipy| numerical processing libraries.
The complete source code can be obtained as a git archive available at  http://birds.eecs.umich.edu/crawler-recovery.git.

We took the configuration space $Q_R$ of the robot to be $(x,y,\theta_0, \theta_1, \dots, \theta_6) \in \mathsf{SE}(2) \times \mathbb{T}^6 = G \times S$.
This comprises the center-of-mass position $(x,y)$ and orientation $\theta_0$ within the plane, which we collectively denote with $g := (x, y, \theta_0)$, and the six joint angles $\theta_j$, $j=1\ldots 6$, which we collectively refer to with $\theta_V := \left(\theta_1, \dots, \theta_6 \right)$.
The latter are intrinsic, i.e. relative to the body and symmetric under translation and rotation of the center-of-mass; they are thus shape variables.

Our robot moved by attaching to the plane at the two locations $l_1$ and $l_2$ with freely rotating pivots, dragging itself with its limbs.
Let us consider the problem of preserving this body motion when the two leg attachment points $l_1$ and $l_2$ are fixed, but a joint motor jams.

The robot is a kinematic system, so that the positions $l_1$ and $l_2$ and angles $\theta_j$ jointly define holonomic constraints which are the native physical constraints $\Omega_P$ on $Q_R$.
Using complex numbers to represent the plane, the following equations compute the endpoints of the limbs $f_1$ and $f_2$ as a function of $(x,y,\theta,\theta_1,\dots,\theta_6)$:
\begin{align}
f_1 := x+iy+e^{i\theta}\left(h_1+e^{i\theta_1}(1+e^{i\theta_2}(1+e^{i\theta_3}))\right)\label{eqn:crawler-anchor-constraints1}\\
f_2 := x+iy+e^{i\theta}\left(h_2+e^{i\theta_4}(1+e^{i\theta_5}(1+e^{i\theta_6}))\right)\label{eqn:crawler-anchor-constraints2}
\end{align}
The robot is subject to the (holonomic) constraints $\frac{d}{dt}f_1=0$ and $\frac{d}{dt}f_2=0$.
These are four constraints on $Q_R$ that make up $\Omega_P$, with $\gamma_P = 0$ (we treat the real and complex components as separate equations).

We arbitrarily chose the parameters $l_1,l_2,h_1$, and $h_2$ to be as shown in Table. \ref{table:crawlers-params}.
These choices roughly match the proportions in Fig. \ref{fig:crawler-sch}.
\begin{table}[h]
	\caption{Crawler Parameters}
	\begin{center}\label{table:crawlers-params}
		\begin{tabular}{l l }
			\hline
			\hline
			Parameter & Value  \\ %\hline
			$l_1$ & $+2.5+i2$ \\ %\hline
			$l_2$ & $-2.5+i2$\\ %\hline
			$h_1$ & $ +1 $\\ %\hline
			$h_2$ & $-1$ \\ %\hline
		\end{tabular}
	\end{center}
\end{table}

\begin{figure}[ht]
	\centering
	\begin{subfigure}[t]{0.45\textwidth}
		\centering
		\includegraphics[width=.925\linewidth]{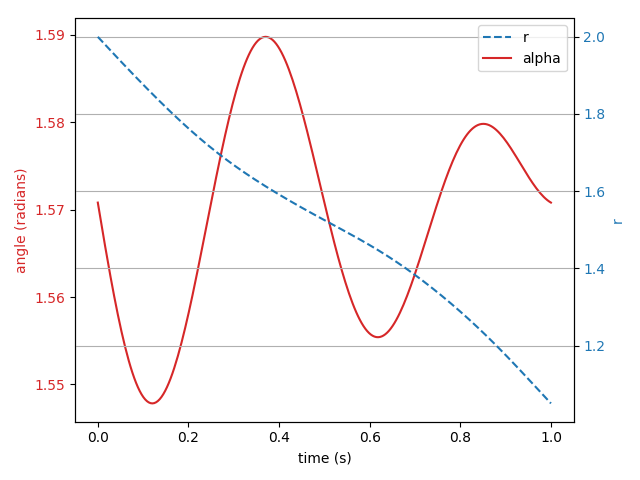}%{crawler_shape_ref.png}
		\caption{$r$ and $\alpha$ coordinates of the template along the $\theta_{orig}$ curve, i.e. the desired reference trajectory on the template.}
		\label{fig:crawler-shape-ref}
	\end{subfigure}\hfill
	\begin{subfigure}[t]{0.45\textwidth}
		\centering
		\includegraphics[width=\linewidth]{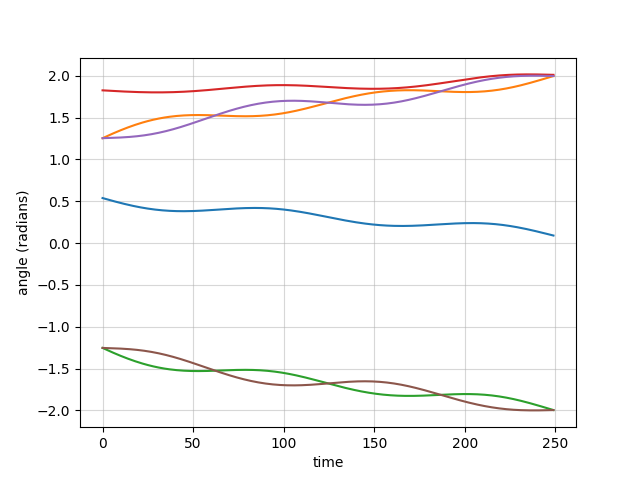}%{crawler-theta-org.png}
		\caption{The six joint angles of $\theta_{orig}$ corresponding to $x_0(t))$}
		\label{fig:crawler-anchor-ref}
	\end{subfigure}
\end{figure}

Our desired motion is depicted in Fig. \ref{fig:crawler-tracking}.
The traces labeled ``desired'' are the nominal inputs without damage.
This roughly corresponds to a serpentine pattern from the CoM, and it is this $g(t), t \in [0,1]$ we wish to maintain.
The required inputs $\theta_j(t)$, which we jointly refer to as $\theta_{orig}$, are shown in radians in Fig. \ref{fig:crawler-shape-ref}.
In the notation of section \ref{sec:math-background}, $(g(t), \theta_{orig}(t))$ is the nominal $x_0(t) \in Q_R$.

We now define a choice of encoding template $Q_E$.
Let the point $q \in \Complex$ be the midpoint of the two foot locations $l_1$ and $l_2$.
We define the encoding template $Q_E = \SE(2) \times \mathbb{R}\times \Sunit$ as the body frame, with additional $(r,\alpha) \in \mathbb{R} \times \Sunit$, as shown in red in Fig. \ref{fig:crawler-sch}.
The template shape coordinates are the distance $r$ of the CoM to the point $q$, while $\alpha$ is the angle of the robot body with w.r.t line $q-(x,y)$ which we encode via \eqref{eqn:crawler-template-basis}.
\begin{equation}\label{eqn:crawler-template-basis}
r\exp(i(\theta_0+\alpha))+x+iy = q
\end{equation}

This definition of encoding template defines a map $(x,y,\theta,r,\alpha) = \phi(\theta_1,\dots,\theta_6)$, which expresses the notion that while we cared about the $\SE(2)$ location of the body of the robot, and the relative location of the robot to $q$, we did not care about specific actuator angles except inasmuch as they influenced those outputs.
Solving \eqref{eqn:crawler-template-basis} for $r$ and $\alpha$ in terms of the $\theta_i$, $h_1$, $h_2$, $l_1$, and $l_2$, we obtain $\phi$.
Again, we point out that there are two independent equations determined by \eqref{eqn:crawler-template-basis}, as we solve the real and imaginary parts separately.
Note that this implies $\phi$ is the identity map on the $(x,y,\theta)$ coordinates representing the robot body in $\SE(2)$.
We evaluated $\phi$ along our original trajectory, yielding the desired output $y(t)$ in the encoding template.
We denoted the resultant shape component of  $y(t)$ by $(r_0(t),\alpha_0(t)) \in \mathbb{R}\times S^1$ (see Fig. \ref{fig:crawler-shape-ref}).

For the encoding procedure to fully define a desireable motion, we needed there to be a unique velocity $(\dot{x},\dot{y},\dot{\theta})$ determined by each pair $(\dot{r},\dot{\alpha})$.
By directly differentiating \eqref{eqn:crawler-template-basis}, we obtained two Pfaffin constraints $\omega_1$ and $\omega_2$, that relate $(\dot{r},\dot{\alpha})$ to $\dot{g}$, with $\gamma_D^1 = \gamma_D^2 = 0$.
Additionally, our choice of $\phi$ defined two equations, which when differentiated, yielded the template constraints $\omega_3 := \exd\alpha$, $\gamma_D^3 := \frac{d}{dt}\alpha_0(t)$, $\omega_4 := \exd r$, and $\gamma_D^4 := \frac{d}{dt}r_0(t)$.
To match the dimension of $G$, a final independent equation was necessary.

As our designed behavior specification, we chose the constraint
\begin{equation}\label{eqn:crawler-extended-constraint}
\omega_5 := \exd x - \exd \theta; ~~~ \gamma_D^5 := 0
\end{equation}
Since this constraint is only on group variables, which are unchanged by $\phi$, this constraint is unchanged when pulled back to $Q_R$ as $\Omega_D^5 := \exd x - \exd \theta$.
This virtual constraint augments the holonomic constraints of $\omega_i$ to generate three equations that define $\Omega_D$ so that \eqref{eqn:reconstruction} is solvable.
This last, arbitrarily chosen $\Omega_5$, is a design choice -- i.e. in $\Omega_D$.
We could have equally used an example motion and a learned constraint $\Omega_L$.

We now assume that the $\theta_1$ actuator is jammed.
We expressed this by adding a differential constraint $\exd\theta_1$ to $\Omega_P$ whose value is identically $0$.
The addition of this constraint modifies $\Omega_P$ to $\tilde{\Omega}_P$, which is one row longer.
If we did nothing, and simply played back the un-jammed components of $\theta_{orig}$ with the $\theta_1$ actuator stuck, we would obtain considerable error in the encoded $(r,\alpha,x,y,\theta)$ motion.
In Fig. \ref{fig:crawler-tracking} the curve labeled ``old'' illustrates the resulting velocity of the CoM should ``playback'' be attempted without some form of recovery.

We thus employ the constraints $\omega_i$ we determined above.
Pulling these constraints back to $Q_R$ defines $\Omega_D$, and combining them with $\Omega_P$ we obtain a six-dimensional non-autonomous differential equation that can be integrated to generate a desired motion. 
For complete details, see Appendix E.
While a six dimensional system may initially seem contradictory, the jamming constraint is very clearly integrable, the six dimensional equation is evolving on a five-dimensional submanifold, which gives us the required $\theta_2,\dots,\theta_6$ we desire.

The solution of this equation is the $\theta_V$ that generates the same desired COM motion despite the seized limb, if such a solution exists.
Numerically integrating, we obtained a new $\theta_V^{rec}(t)$, $r(t)$, and $\alpha(t)$, shown in Figs. \ref{fig:crawler-shape-ref} and \ref{fig:crawler-anchor-new}.
Note that we cite Fig. \ref{fig:crawler-shape-ref} for the new encoding template curve as well;
this is intentional, as the new and old encoding template curves are numerically identical.
\begin{figure}[ht]
	\centering
	\includegraphics[width=.6\linewidth]{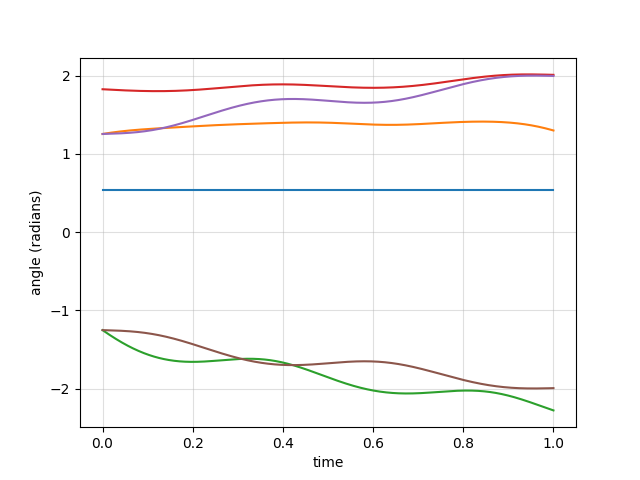}%{crawler_theta_new.png}
     \caption{The recovered joint angles $\theta_V^{rec}(t)$. As expected, with a jammed joint, one angle is constant. By resolving for a new set of limb angles that satisfy the constraints, the template motion is recovered}	\label{fig:crawler-anchor-new}
     \end{figure}
%\begin{figure}[ht]
%	\centering
%	\begin{subfigure}[t]{0.45\linewidth}
%	\includegraphics[width=\linewidth]{gcoun5}%{crawler_theta_new.png}
%	\caption{}
%	\label{fig:crawler-anchor-new}
%    \end{subfigure}
%    \begin{subfigure}[t]{0.47\linewidth}
%        \hspace{1.5cm}
%    	\fontsize{25pt}{4em}{
%    	\resizebox{75mm}{!}{ \input{figures/connection.pdf_tex}}}
%      
%    	\caption{}
%    \end{subfigure}
%    \caption{ (left) The recovered joint angles $\theta_V^{rec}(t)$. As expected, with a jammed joint, one angle is constant. By resolving for a new set of limb angles that satisfy the constraints, the template motion is recovered. (right) A schematic representation. The anchor (upstairs) and the template (downstairs) are related by projection $\varphi$.  The differential section $D\varphi^{\dagger}$ must be chosen to lie in the affine subspace $\Omega^{-1}{\gamma}$, which is free  in $\text{ker} \Omega$. Curves $x(t)$ and $\tilde{x}(t)$ that satisfy the constraints both project to $y(t)$.}
%\end{figure}

The performance of our recovered joint inputs is shown in Fig. \ref{fig:crawler-tracking} as the ``recovered'' trace.
It appears to recover the desired CoM velocity (and thus, group position) to within visible accuracy.
This is especially notable in contrast to no recovery at all (the ``old'' trace in Fig. \ref{fig:crawler-tracking}).

\subsection{Hexapod robot}\label{sec:Enepod}
\begin{figure*}[!ht]
	\centering
	\includegraphics[width=\linewidth]{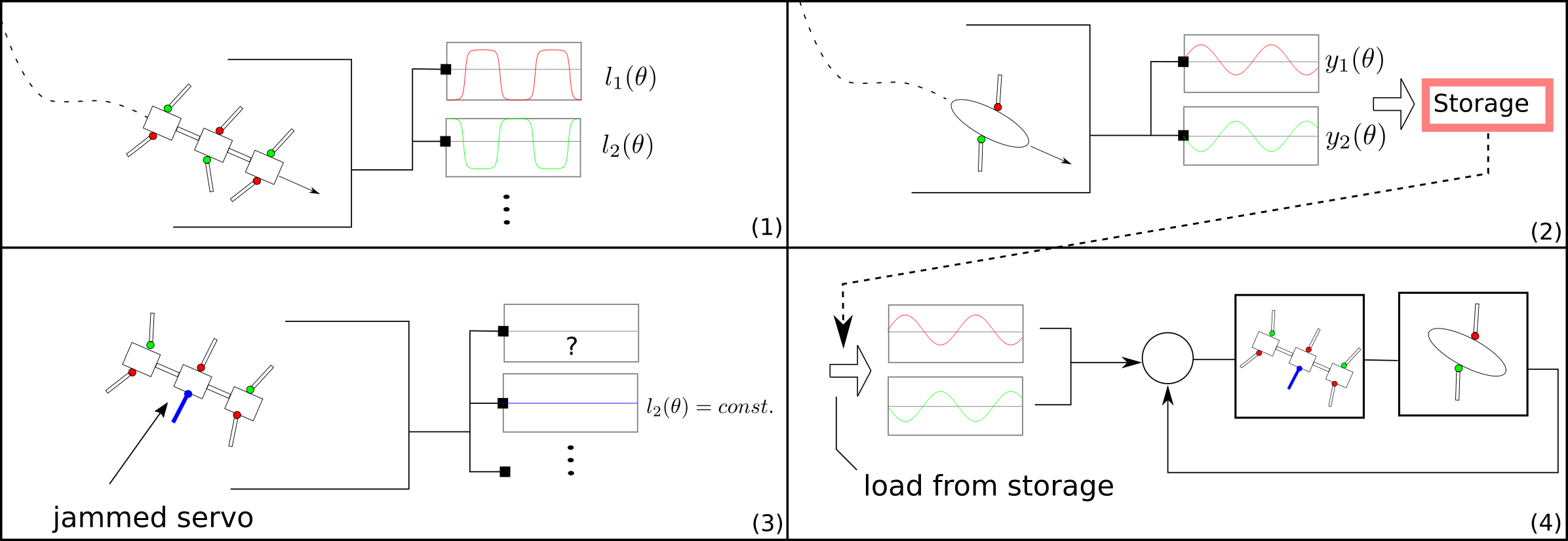}%{big_fig.png}
	\caption{Overview of the fast recovery process. %
		\textbf{(Training)} Assume an existing desired behavior $x_0(t)$ is found for the robot (here a hexapod; (1)) using control $u(t)$, e.g. via costly optimization. %
		It satisfies both physical $\Omega_P$ and design $\Omega_D$ constraints: $\Omega_{P,D} \cdot {\dot x}_0 = \gamma_{P,D}$.
		Project its state down to a candidate encoding template by identifying a collection of observations sufficient for representing the desired behavior (here a sprawled biped; (2)), thereby selecting $\phi$.
		Selecting further a collection of learned constraint forms $\omega$ on this template. %
		Learn and record the $\eta(t) := \omega(\phi(x_0)) \D \phi {\dot x}_0$ of \eqref{eqn:x0} in memory, and store the pullback of $\omega$, the learned constraints $\Omega_L := \omega \circ \phi \cdot \D \phi$. %
		\textbf{(Recovery)} Robot has failed (blue leg has jammed motor in (3)); it has a new physical $\Omega^*_P$. %
		If the new $\Omega^*_P$ is known, pick a subset $\Omega^*_L$ of $\Omega_L$ such that together $\Omega^*_P, \Omega_D, \Omega^*_L$ are exactly full rank, and solve for a new $\dot x$ which will recover the behavior (e.g. in (4)). %
		If $\Omega^*_P$ is not known, iteratively learn a new control $u^*$ minimizing the constraint violation cost $J[x^*,u^*]$ of \eqref{eqn:con-cost-function}. %
		This cost function admits $x_0$ as an optimum if it is achievable, and all its optima have the same encoding template motions as $x_0$ (e.g. in (4)). %
		Since the cost function has a richer gradient, the optimization can converge much faster than the naive optimization that created $x_0$ in (1). }
	\label{fig:iteration}
\end{figure*}
Since we are ultimately interested in recovery on robots with truly unknown (or least, unmodeled) dynamics, we continue our examples with a physical mechanism. 
As the first step , we need an appropriate encoding template.
For biomechanists, the difference between ``running'' and ``walking'' is defined in terms of the energy reservoirs participating in the exchange generating the motion.
In walking, potential energy exchanges with kinetic energy by vaulting over a rigid leg; thus ground speed is lowest when the center-of-mass is highest.
In running, elastic energy of stretched tendons and muscles exchanges with kinetic energy; thus ground speed is highest when the center-of-mass is highest.
Thus, the kind of gait appearing (running vs. walking) can be encoded in terms of total energy in these reservoirs.
We designed a six-legged robot to facilitate the measurement of elastic energy storage in its legs.
This, we hoped, would allow us to define an encoding template in terms of these energy exchanges, and test our strategy on a physical device.
Simulation study of recovery on a low-dimensional dynamic running device further supported the notion that energy was important for motion. 
Appendix B has a complete description of this study, which is omitted here for brevity. 
Additionally, while high-fidelity model of this hexapod might include discontinuous impacts, which would seem to imperil the smooth analysis posed above, the alternating tripod gaits the robot is restricted to has a differentiable approximation that admits our smooth analysis \cite{kelly1995geometric}.

\begin{figure}[ht]
	\begin{subfigure}[t]{0.45\textwidth}
	\centering
	\includegraphics[width=\linewidth]{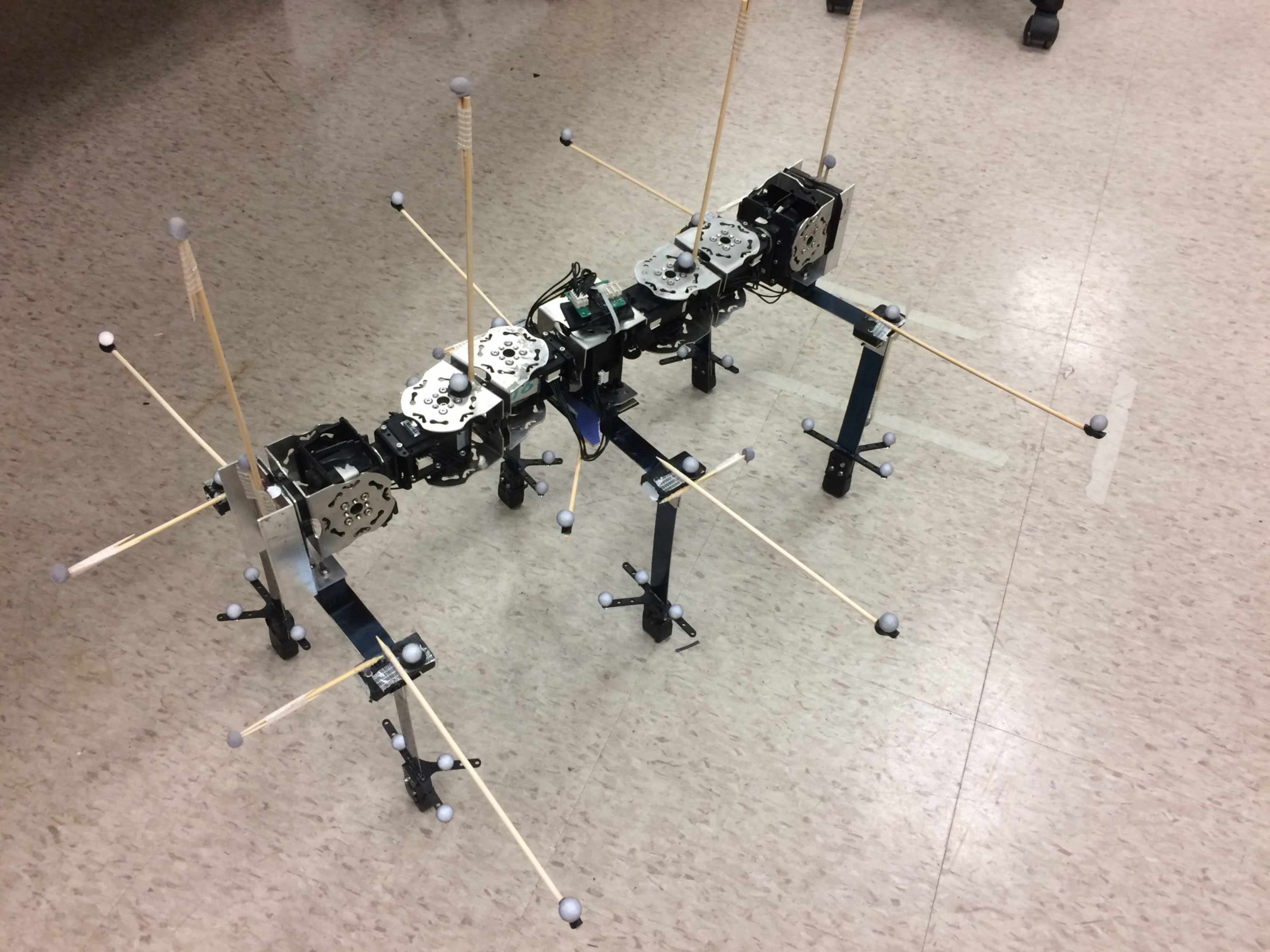}
	\caption{ } \label{fig:Enepod}
	\end{subfigure}
   \begin{subfigure}[t]{0.45\textwidth}
   	    	\fontsize{14pt}{4em}{
   		\resizebox{80mm}{!}{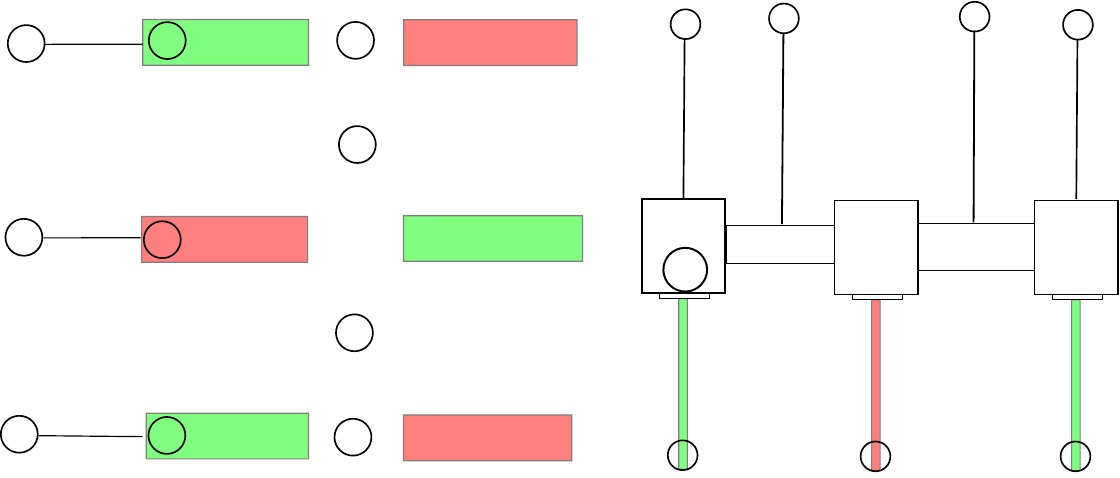}}
\caption{}\label{fig:Enepod-coords}
\end{subfigure}
\caption{(a) The Enepod robot. %
	We mounted retroreflective markers on wooden spokes, increasing our resolution for measuring deflections of the legs in real time using the marker motion tracking system. (b) Leg groupings on the Enepod. %
	The grouping (BL,MR,FL; in green) defines the \concept{left tripod} of legs, which move in phase which each other, and antiphase to the \concept{right tripod} (BR,ML,FR; in red). %
	This grouping of legs was used for both horizontal and vertical springs. %
	Some markers have been omitted for visual clarity.}
\end{figure}

The robot (``Enepod'') is depicted in Fig. \ref{fig:Enepod}.
It consisted of a chain of 7 motor modules (Robotis Dynamixel EX106 and MX64) as an actuated backbone, connected to six passive spring steel ($1/16"$ \#1075) legs.
The legs were mounted to the EX modules, as they provide more torque.
The springs were flexible enough to exhibit deflections of more than $1\,\text{cm}$ at the foot during motion making it feasible to sense their deflections using a motion tracking system.
This provided an instantaneous window into the elastic energy stored in the body at any given time.

We generated a robot gait as a sequence of timed position commands which were carried out by individual control loops in the motor modules.
The gait we chose was an \concept{alternating tripod} gait analogous to that used in the RHex hexapod \cite{xrhexkod2010}.
In this gait the feet are grouped into two collections of three feet (\concept{tripod}).
Feet in a tripod moved in phase with each other, and anti-phase with the feet of the other tripod.
If the system were perfectly rigid, each tripod would be undergoing an identical motion.
With elastic legs, even though they receive the same commands, the dynamics of the body and of contacts alter the response.
Such flexible limbs would have been a major challenge for a dynamic model.
However, since our method does not need a predictive model we did not encounter this difficulty.

As is appropriate for a periodic gait, we analyzed the motion of the robot with respect to a kinematic phase estimate \cite{sharbafi2017bioinspired} obtained using the tool \emph{Phaser} \cite{phaserRevzenGuck2008} from motion tracking data collected with a retroreflective tracking system (Qualisys; with 10 Opus-310 cameras at 250 FPS; Qualisys Track Manager v2.17 software interfaced to custom python SciPy 1.0.0 code using the Realtime API v1.2.

The space of gaits we considered is spanned by the parameter space $\mu \in \left[-1,1\right]^5$.
The values of $\mu$ determined the signal that drives the center module $MC$ (see Fig. \ref{fig:Enepod-coords}).
The other six modules input signal remained unchanged.
We compared convergence rates for optimizing $\mu$ with respect to a conventional cost function to optimizing $\mu$ with respect to a learned behavior specification $\Omega_L$.

We performed this process in two stages.
In the first stage, we manually designed a gait that achieved forward translation.
Using the notation used in \S \ref{sec:math-background}, this nominal gait is $x_0(t)$.
We then evaluated our chosen output functions -- chosen for being obvious stand-ins for energy reserviors -- along this cycle.
We constructed a representation of the functions' values as a Fourier series in phase.
Using this model, we differentiated to obtain the necessary $\omega_i$, and $\eta_i$.
By pulling back these functions to the original state-space using our numerical estimate of phase, we obtain $(\phi, \Omega_L, \gamma_L)$ --- the \concept{learned constraints} that we introduced in \S \ref{sec:recovery-constraints}.
We used these $\Omega_L$ to define a cost function $J$, exactly as in \eqref{eqn:con-cost-function}.

\subsubsection{Encoding Template}
We defined four output functions whose images, together with phase, comprise the encoding template.
These outputs were the vertical and horizontal deflections of each of the two tripods.
The intuition behind this choice of outputs was that: (1) the tripods act independently; (2) the legs in a tripod can trade off each other; (3) the vertical and horizonal bending of the legs was, by design, independently taken up by different springs; (4) vertical and horizontal bending is expected to occur at different phases.
Thus each tripod had two elastic energy reservoirs, one each for horizontal and vertical deflections, expected to act at different phases.
This can easily be seen from the horizontal and vertical projections of the classical Spring Loaded Inverted Pendulum model \cite{blickhan1989}.
The phase dependent changes in these tripod-average deflections constituted the learned constraints $\Omega_L$ that we employed in lieu of any other model information to characterize our desired motion and produce a behavior specification violation cost, exactly as described in \eqref{eqn:con-cost-function}.

We calculated the vertical spring deflection $V_i, i=1,2 $ from marker locations, coding it as an angle rather than as a linear displacement.
This angle, between the horizontally orientated spring-steel members and the central spine, is monotonically related elastic energy stored (according to e.g. beam theory), was easy to measure given our instrumentation, and we found it empirically to vary in a periodic manner.
We collected six distinct vertical deflection angles at every time-step despite the left and right angles resulting from the deflection of the same double-sided leg, because each leg was  clamped to the body in its middle, allowing each side of a leg to bend independently in the vertical direction.
Fig \ref{fig:vert-def} depicts an example time series collected from the Enepod striding forward with its limbs cycling at $2\,\text{Hz}$.

\begin{figure}[ht]
	\centering
	%\begin{subfigure}{0.45\textwidth}
	\begin{subfigure}[t]{0.45\textwidth}
	\centering
	\includegraphics[width=\linewidth]{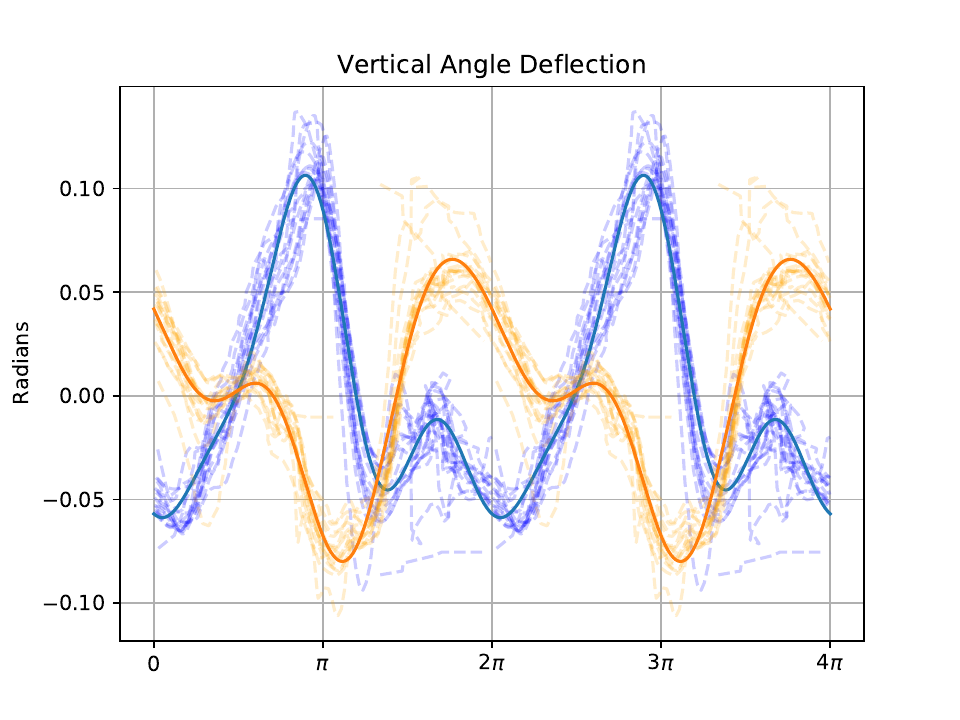}%{figures/VERT.pdf}
	\caption{Summed vertical deflection signals.}\label{fig:vert-def}
	\end{subfigure}
    %\begin{subfigure}{0.45\textwidth}
   \begin{subfigure}[t]{0.45\textwidth}
	\centering
	\includegraphics[width=\linewidth]{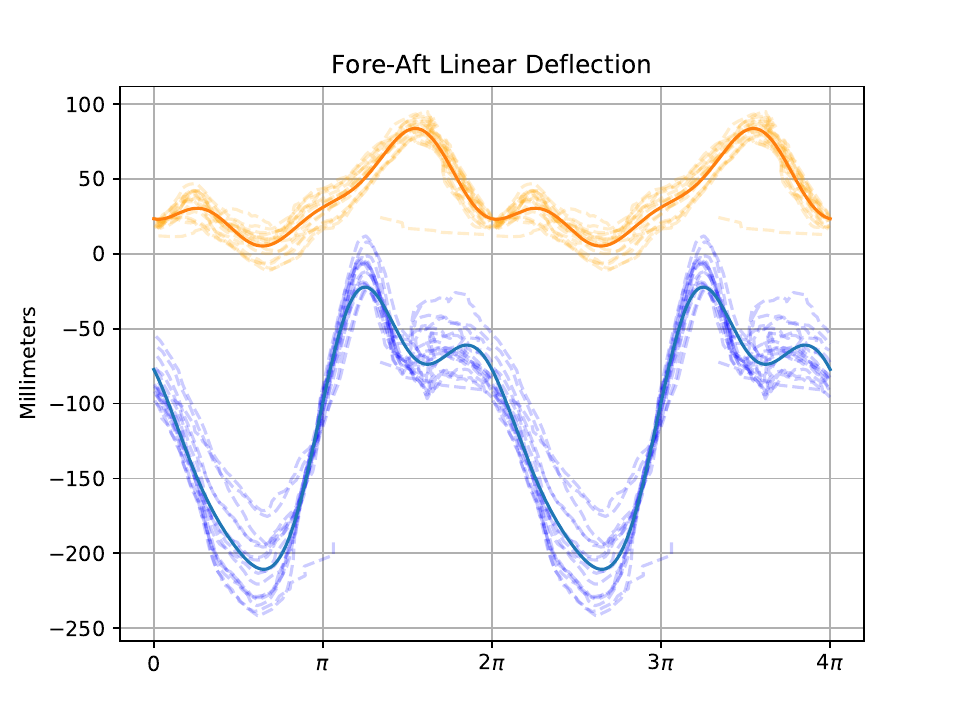}%{FA.pdf}
	\caption{fore-aft deflection signals. %
		}\label{fig:Enepod-hori-def}. %

	\end{subfigure}
\caption{The first two principal components of tripods $H_1$ and $H_2$, respectively. We plotted their values as function of phase in individual strides (dashed lines), and the Fourier-series fit of order 3 to $n=15$ strides (solid lines)}
\end{figure}
We measured the fore-aft deflection of the vertically-mounted springs (e.g, the deflection in the horizontal direction) differently from the vertical deflection.
We used the marker sets indicated in Fig. \ref{fig:Enepod} to define the two centroids - $C_{top}^i$, and $C_{foot}^i$, respectively, where $i$ indexes the leg.
We projected these two centroids into the $(x,y)$ plane, and performed a principal component analysis \cite{pearson1901liii}.
The first two principal component projections, taken as a function of phase are the two horizontal outputs $H_1$ and $H_2$ (see Fig. \ref{fig:Enepod-hori-def}).
Physically, there are two principal axes of horizontal deflection, representing two elastic energy reserviors associated with horizontal bending, but it turned out that these axes do not correspond to bending parallel to the major line of the central spine.
This suggests that the two reserviors are approximately $90^\circ$ out of phase.

\subsubsection{Goal function}\label{sec:Enepod-goal}

We built a nominal value for each output function by measuring its value while the robot was executing the nominal gait ($x_0(t)$).
By computing and caching Fourier-series approximations of the output functions, we defined the constraints.
As can be seen in Fig.  \ref{fig:Enepod-hori-def} and Fig. \ref{fig:vert-def}, these output functions are strongly periodic and lack apparent spectral complexity, allowing an order $4$ Fourier series to fit them well within measurement noise.
For the disrupted robot, we used these Fourier series to define the cost function as shown in \eqref{eqn:con-cost-function}.

As the robot is not perfectly periodic, to evaluate the cost function for each choice of parameters, we used a windowed average of the goal functional over windows of $n=35$ strides (cycles)
We fixed the gait frequency at $2.5\,\text{Hz}$, giving a duration of $t_f = 35 \times \frac{1}{2.5}=14\,\text{sec}$ for each cost function evaluation.

\subsubsection{Optimization results}

We conducted the optimization using the Nelder-Mead algorithm on the $\mu$ parameters that define the signal driving the central module.

\begin{figure}
	\centering
	\begin{subfigure}[t]{0.4\textwidth}
	\centering
	\includegraphics[width=\linewidth]{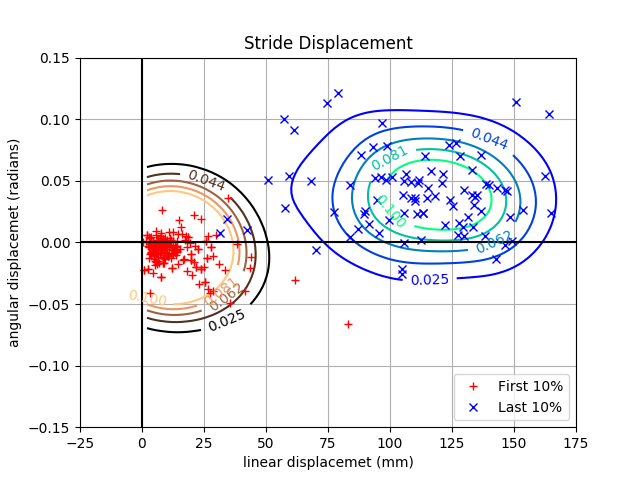}%{enepod-performance.png}
	\caption{}\label{fig:enepod-results}
    \end{subfigure}
   \begin{subfigure}[t]{0.4\textwidth}
	\centering
	\includegraphics[width=\linewidth]{gcoun12}
	\caption{}\label{fig:enepod-cost-result}
     \end{subfigure}
    \caption{ (a, left) A comparison of pre and post optimization.
    We represented $\xi = (\xi_x, \xi_y, \xi_{\theta})$ by the 2D position with  $x=\left(\xi_x^2+\xi_y^3\right)^{1/2}$ and $y=\xi_\theta \in [0,2\pi]$. %
	We plotted the results of individual strides (first $10\%$ of strides red ``+''; last $10\%$ blue ``x'') and the contours of a kernel smoothed density produced from those data (black to orange contours for first $10\%$, blue to cyan for last $10\%$). %
	(b, right) Performance of the cost function by iterate --  absolute cost at the $i$-th iterate (dots, black), and the best-cost-so-far (``x'' marks, teal). %
	}
\end{figure}

The evolution of the cost is displayed in Fig. \ref{fig:enepod-cost-result}.
We terminated the optimization after we had approximately reduced the cost by $40\%$.
Our choice for this was empirically motivated: the robot had achieved acceptable performance, as depicted in Fig. \ref{fig:enepod-results}.
This termination condition is fairly arbitrary, and future work will include a more principled approach for determining termination.
After $n=36$ iterations, where a single iteration corresponds to $35\pm 1$ cycles executed at a set of hyper-parameters $\mu$, the mean displacement per stride was significantly enhanced.

The distribution of motion outcomes at the end of optimization is wider (Fig. \ref{fig:enepod-results}), but one should keep in mind that Nelder-Mead inspects some poor parameter choices in iteration 26 and 33.

The key result is that with $N=36$ iterations, without a predictive model of the robot, and without directly measuring the robot's motion in the plane --- which is the actual goal for the original gait optimization --- we were able to regenerate useful forward motion which is nearly as effective as the initial optimization.
All this was done by enforcing constraints that were learned by measuring the working robot.

\section{Discussion}

We have shown that by using our newly defined notion of a \concept{behavior specification} and \concept{encoding template} we can not only represent a broad class of existing physical dynamics and control problems, but also model an important class of failures --- namely the loss or addition of constraints --- as a deletion or insertion of rows in this specification.
Given the problem of recovering a robot behavior after such a failure, we have shown solutions for two cases.
When the post-failure specification is known, we demostrated a closed form solution (see \ref{sec:crawler}).
When the post-failure specification is not known, we have shown how a reasonable choice of virtual constraints learned by observing the encoding template projection of the desired behavior can be used to rapidly (re-)learn an equivalently desireable behavior (see \ref{sec:Enepod}).

The core property we exploit is a trivial feature of linear equations:
if there are $m$ equations in $n$ variables, and $n > m$, there is a specific solution $v_s$, and a nullspace of values $v_n$ such that $v_s + v_n$ is still a solution.

By keeping track of the linear/affine differential constraints which comprise the behavior specification, we preserve the ability to exploit the nullspace of velocities that preserve our behavior specification as the system changes.
The addition or removal of constraints drops or activates low priority constraints, buffering our scheme against the violation of the design constraints $\Omega_D$.
If through control we can force a damaged system to move within the desired nullspace the resulting behavior will be identical, and any solution that exactly restores the desired behavior must be of this form.
Even if we cannot force the constraint violation cost function \eqref{eqn:con-cost-function} to zero, using it brings us closer to a desireable behavior while being neutral to changes that have no effect on the desired outcomes.

\subsection{Specialization of the result to locomotion}

Robots designed for locomotion often have a natural partition of their configuration space \cite{ostrowski1998geometric}: the body frame evolves in $\SE(3)$ or a subgroup thereof, while other, more arbitrary dynamics account for the motions of the limbs (more generally, variables which determine the shape of the robot).
This natural splitting of variables exists in all systems that are symmetric under a Lie group of transformations.
Such a configuration space is locally of the form $Q=S \times G$ where $S$ is the shape space and $G$ is a group representing the symmetry of the environment, typically sub-group of $\SE(3)$.
The interaction between the body-frame, body, and world is mediated by constraints that differentially relate body-frame velocities to limb velocities.
Usage of so-called \concept{principal G connections} in this context has demonstrated that the our required structure is present; indeed, if the constraints are symmetric under $G$,  \eqref{eqn:reconstruction} is little more than a \concept{non-holonomic connection} \cite{bloch1996nonholonomic} expressed in coordinates.
Our construction is effectively postulating that we can assign a connection to the template, and pull this connection back to the anchor.

It is often convenient to write models such that: (1) the space $Q_R = S_R \times G$, and that (2) the map $\varphi = (\varphi^S, \text{id}_G)$ is identity on the group $G$, and $\varphi^S : S_R \to S_E$, for $Q_E = S_E \times G$.
This assumption corresponds to identifying an encoding template $Q_E$ that has the same body-frame (usually taken to be the center-of-mass frame) as the full robot.
For example, if we had a six-legged robot that moved in the plane, we could have an encoding template that was a kinematic car whose center-of-mass and body axes coincided with the robot's for all time.
Then, by recovering the chosen trajectory of the kinematic car, the group motion of the anchor robot is also preserved, even though the robot's limb motions may be very different from their original behavior.

A helpful advantage of a system that is symmetric under a group $G$  is that the dimension of $G$ is a known constant.
Thus, we could immediately obtain the number of required constraints necessary to determine a desired $g(t)$.
If the group has dimension $d$, we necessarily and sufficiently require that we have $d$ linearly independent constraints in the behavior specification.
When this condition is satisfied, $g(t)$ is uniquely determined by a given $s(t)$.

\subsection{On the choice of output functions}\label{sec:obs-funs}
Unlike the classical approaches of mechanical modeling, we point out that the behavior specification constraints do not need to be direct descriptions of constraint forces attributed to the robot's mechanical structure.
The constraints can be written in terms of outputs evaluated on examples of the desired behaviors -- differentiated, they define the $\Omega_L$ constraints that are required to be satisfied.

At the extreme, recall the behavior specification of a reference trajectory: the constraint forms are $\omega_i := \exd y^i$ and we can use the value constraints $\eta_i(t) := \frac{d}{dt} y_j(t)$.
If $\dim Q_E$ such constraints were defined, this is merely a representation of a desired reference $y(t)$ to be tracked, given a fixed initial condition $y(0)$.
However, these constraints need not be expressed in terms of the native coordinates $y_i$ of $Q_E$.
Another version of this same argument is when the $\omega_i$ are the derivatives of \concept{output functions} $f_j : Q_E \to \mathbb{R}$.
As long as the collection of derivatives $\nabla f_j$ is full rank, the chain rule allows constraints to be synthesized just as effectively in terms of the $f_j$ -- an arbitrary choice of output variables $y$ -- allowing those to be selected for convenience of measurement.

We both emphasise, and have demonstrated, that experimental data is completely adequate to build such constraints, as long as data is sufficiently rich to support a suitable collection of locally defined output functions.
To define $\eta$, the robot can execute its desired motion, and the time-series that results from evaluating the output functions \emph{is} $\eta$ by definition.
The recovery problem is to restore this time series via control.
The usage of output functions in such a manner mitigates the need to develop any predictive model, such as the dynamics of the robot, which is generally a challenging task.

The condition of sufficient rank is essential to our process, and likely translates in practice to a need for a sufficient rank and a good condition number.
Since by assumption the dimension of $Q_R$ is greater than that of $Q_E$, it can be shown with straightforward transversality arguments (e.g, Chapter 2 of \cite{golubitsky2012stable}) that the set of output functions that are adequate to meet our necessary rank condition is an open and dense set. 
For precise details, see Appendix C.
Intuitively, its unlikely that randomly selected vectors are linearly dependent in a high-dimension space, and very rectangular random matrices will generally have good condition numbers.

More generally, the differential constraints themselves can be considered \concept{output functions}, only that their domain of definition is $\T Q_E$, rather than $Q_E$.
For any differential form $\omega$, it can always be evaluated on a given observed output curve to return a time sequence of values $\eta$.
Thus, we need not be restricted to $\omega$ choices that arise as exterior derivatives, i.e. non-exact forms can also be used.
A designer is free to employ real-valued functions of $Q_E$ and use their exterior derivatives, or to directly define linear functions of velocity, i.e. differential forms, on $\T Q_E$.

\subsection{Behavior Specifications can represent kinematic synergies}

In biomechanics the concept of \concept{kinematic synergies} is used to represent the observation that animal muscle actuation is often coordinated in such a way that motions occupy low dimensional subspaces of the space of possible actuation combinations (e.g. human manipulation tasks  \cite{jarque2019kinematic, daffertshofer2004pca, jarque2016using}).

From a formal mathematical standpoint, collections of behavioral constraints are unrelated to coordinated motions, because coordinated motions are subspaces of the tangent space, and constraints define subspaces of the cotangent space.
However, if we are able to equip the anchor space with a metric, we can dualize the constraints (via the musical isomorphism \cite{lee2013smooth}) to interpret them as defining a control distribution, including those that represent synnergies.

\subsection{Scaffolding for learning movement quickly}

Humans exhibit a series of developmental milestones while learning to walk \cite{adolph2013road}.
Simulation studies (see \cite{bongard2011morphological}, and the references therein) and common practice by roboticists has shown that incorporating optimization milestones into a scaffold of nested behaviors can dramatically improve the rate at which robots can learn complex physical behaviors.

Since the pullback of a differential form is defined for any full-rank map between manifolds, our approach suggests a natural extension to a scaffold, i.e., we can just as easily have a sequence of encoding templates $Q_{E_1}, \dots Q_{E_n}$ (with the convention $E_0 := Q_R$) related by projections where $\phi_{n-1:n}$ maps template $E_{n-1}$ into $E_n$.
This gives rise to a corresponding chain of nested behaviorial specifications $(\phi_{n-1:n}, \Omega_{E_n}, \gamma_{E_n})$.

A scaffold like this, were constraints are iteratively pulled back, allows a learning strategy to construct complex behaviors for highly actuated robots out of lower-DoF \concept{proto-behaviors}.
Designers, be they engineers or autonomous optimization tools, could initially design a curve $x(t) \subset E_{n-1}$ that obeys a behavior specification for $E_n$.
Then, they can use $x(t)$ to augment an existing behavior specification on $E_{n-2}$.
In designing a new curve in $E_{n-2}$, we gain the ability to both \emph{preserve} the first behavior, while enforcing a new one.
As $n$ decreases, the dimension of the corresponding encoding template grows, allowing us to constructively lift low-dimension component behaviors into increasingly complex anchors.

\subsection{Behavior Specifications are not a planning tool}

It is common in contemporary robotics to plan the motions of a complex robot (e.g. a car) using a simplified representation (e.g. a unicycle model template).
Once a plan is created which meets certain feasibility heuristics, the plan is given to an optimizer which computes a detailed actuation schedule for its implementation, based on a detailed anchor model.
With this in mind, a key issue in using templates for planning is having a guarantee that a planned template behavior can be realized by some choice of input to the anchor.

Much of the efficacy of our method for behavior recovery is due to not requiring this property.
A key feature of our encoding templates is that we do not require that all curves in the encoding template can be realized as trajectories of the anchor; we only require that one distinguished, desired $y(t)$ can.
Equivalently, we only insist that the constraints are satisfiable along the specified $y(t)$, rather than everywhere on the encoding template.

It may very well be that the combined $\Omega_P$ and $\Omega_D$ are over-determined other than on $x(t)$, and cannot be simultaneously satisfied.
Thus the constraints $\omega_i$ are not suitable for planning.
For example, if the template constraints corresponded to a kinematic car, the trajectories of the the car other than the one distinguished car motion corresponding to our desired motion are not required to be achievable by the physical car.
If the $\omega_i$ are defined over the entirety of $Q_E$, they could be used to classify multiple output curves as meeting or violating the constraints, but we emphasise that this is not the same as requiring every output curve to be achievable.

\subsection{Relationship to Output Tracking}

A vast literature in control theory is dedicated to the problem of \concept{output tracking} -- producing a desired trajectory of output variables.
While our constraint-based optimization function offers an empirically rapid solution technique for such a problem, our contribution is more accurately reflection in the idea that the constraints $\omega_i$ \emph{define} the output $y(t)$.
We elected the constraints first, and evaluated them as output functions along a known behavior $x_0(t)$, which is how we originally obtained the definition of $y(t)$.

In our formulation, the desired output $y(t)$ is an \emph{encoding} of the desired behavior that was generated to represent it to facilitate recovery.
Part of the novelty we claim for our work is the observation that a desired behavior specification can be obtained from a known $x_0(t)$ in a very cavalier way -- pretty much any set of $\omega$, and any $\phi$ that meet our requirements are equally good for defining $y(t)$.
We contrast this from methods such as hybrid zero dynamics \cite{westervelt2003hybrid,grizzle2008hybrid} , wherein ``virtual constraints'' are fiats defined by the engineer independently of the systems underlying dynamics, which are overridden by control when a set of strict technical conditions is satisfied.
Computational accuracy constraints suggest that having the matrix $\omega$ well conditioned is advantageous, but we have found no other requirement to be of great practical importance.
That $y(t)$ is a derived quantity is part of our motivation for omitting it from the notation $(\phi, \omega, \eta)$.
It also motivates the adjective ``encoding'', in that we have found a representation of a desired behavior, but hints that such encodings are not unique.

\section{Conclusion}

The key contribution of our work here is a novel method for recovering robot behaviors after damage to the robot renders the previous actuation policy ineffective.
We offer a key insight -- although damage is complicated to understand in terms of the changes it causes to a control distribution, many common forms of damage take the form of low-rank changes to the constraints that define the robot dynamics.
Thus, we propose a dual formulation for both dynamics and desired behaviors: \concept{behavior specifications}, which are defined in terms of differential forms and their desired outputs.
This cotangent bundle formulation has a distinct mathematical advantage in that it make it easy to encode example behaviors and recover them when low-rank changes to the constraints occur.

With this approach we have shown large speedups in the ability of physical robots to re-learn a desired behavior, and shown how a simulated system can recover to within numerical precision when the damage model is known.
Our approach has ties to many existing ideas in control, theoretical mechanics, and neuromechanical control in animals.
These suggest to us that the \concept{behavior specifications} we propose here will help connect fields and inform our future work for many years to come.

%\bibliographystyle{siam}
%\bibliography{References}

\section{Appendix}
In this appendix we aim to expand on the technical details of the main text, provide additional motivation, as well as provide a dynamic example. References that start with the prefix ``M'' refer to the relevant reference of the main text, e.g., ``\S M-2.2'' is \S 2.2 of the main text.

\subsection{Encoding Templates Motivation}
We distinguish our approach by taking advantage of two dimension-reducing constructions. 
The first is the \emph{templates and anchors} hypothesis (see \citep{fullTemplates1999} for a detailed introduction). 
Briefly, it asserts that robots with complex models (the \emph{anchor}) have coupled or dependent state variables that behave ``as if'' the robot had a lower-dimensional model (the \emph{template}).
In other words, if $f : M \to \T M$ is the anchor vector field, there exists an attracting invariant immersed submanifold $N \subset M$ (the template), $\dim(N) < \dim(M)$ such that $f\left.\right|_N: N \to \T N$.
Considering the anchor as an invariant submanifold has many desirable properties, but it is also a strong requirement.
A weaker yet simpler approach is to consider the template as a manifold $N$ with vector field $g : N \to \T N$, with submersion $\varphi : M \to N$ such that $g = \varphi_* f$. 
In this regard, we think of the template as a virtual system whose trajectories are the ``shadows'' of those of $f$. 

In this language, the behavior of interest is represented by a solution $\phi^t(r_0) \in N$. e.g. - if a cockroach has a Spring-Loaded Inverted Pendulum (SLIP) as a template, the behavior of the SLIP is what we wish to preserve, in face of damage to the anchor, where the anchor has sufficient control authority to implement a SLIP.

The second construction (detailed in appendix C, below) we take advantage of is offered by geometric mechanics and control.
For lagrangian systems with nonholonomic constraints that are symmetric under a group action, locomotion may be represented by a \emph{connection} on a \emph{principal fiber bundle}. \citep{ostrowski1998geometric, bloch2003nonholonomic,bloch1996nonholonomic}. 
The resulting \emph{reconstruction equation} neatly expresses the motion in the group as a function of internal shape variables.
E.g, the total displacement of a planar robot moving its limbs can be thought of as an element $g \in SE(2)$ resulting from cyclic limb motions $l(t), t \in [0,T]$ - literally, the ``shape'' of the robot is modulated to effect displacement of the center-of-mass.

\subsubsection{Templates and Anchors}\label{sec:temp-anch-review}

If we inside abide  by the model that a template (an encoding template is not a template in this formal sense) is a normally attractive invariant manifold (NAIM) as taken in  \citep{fullTemplates1999}, we can conclude the existence of an important quantity - asymptotic phase.
Asymptotic phase is a concept that has received considerable attention in the dynamics community \citep{fenichel1974asymptotic,alexander1994smooth,hirsch2006invariant,fenichel1977asymptotic}. and we briefly summarize its definition and some essential properties here. 

Let $M$ be a smooth manifold with vector field $f : M \to \T M$ with flow $\phi^t(\cdot)$.
Suppose that the manifold $N \subset M$ is asymptotically stable under $\phi^t(\cdot)$. 
If there exists smooth submersion $P : M \to M, P(M) = N$ such that $\forall x \in M$, and for any $q \in N$, $q \neq P(x)$,  
\begin{equation}\label{eqn:def-phase}
\lim_{t \to \infty} \frac{ \norm{ \phi(x,t) - \phi(P(x),t)}}{ \norm{ \phi(x,t) - \phi(q,t)}} = 0 
\end{equation}
We call $P$ the phase map, and the value $P(x)$ the phase of $x$.
The point $P(x)$ is unique in the sense that $x$ converges with $P(x)$ more rapidly than with any other point putative point $q$.
$P$ is a nonlinear projection, as $P \circ P = P$.
Points that are in-phase have the same asymptotic behavior.
Stable normally hyperbolic manifolds always have asymptotic phase. 
Points that are in phase asymptotically approach each other as $t \to \infty$ - geometrically, the projected point is the ``shadow'' of infinity many points that share the same phase. 

It is this last property that we wish to carry by analogy our simpler case of having the anchor $M$ and the template $N$ be separate manifolds.
Asymptotic phase identifies anchor states with template states -- if $N$ is a NAIM, it also conveys dynamic information.
However, if $\varphi: M \to N$ is merely any submersion, it is a non-linear projection that allows use to \emph{algebraically} relate the state of the template with a family of equivalent states in the anchor, regardless of how the dynamics of the two systems are related.
In the sequel we will employ algebraic phase maps to define \emph{encoding templates}.
Distinct from the regular template above, an encoding template does not have dynamics conjugate to the anchor.
It will function as a collection of \emph{observation variables} that will be shown to fully characterize a desired motion more parsimoniously than the state of the anchor.

In the sequel, we will reserve then notation $\varphi$ for submersions in this context, and will use the terminology ``phase map'' for it, whether or not it is the formal notation of asymptotic phase.
We will use the terminology ``phase of a point'' as the image value under this, e.g. if $\varphi(y) = x$, then $x$ is the ``phase'' of $y$.
We will be explicit about the domain and codomain of such functions to avoid confusion.

\subsection{Example : CT-SLIP }\label{sec:ctslip}

\subsubsection{Problem Statement}

The Clock-Torqued Spring Loaded Inverted Pendulum (CT-SLIP) \citep{seipel2007CTSLIP} model is a hybrid dynamical system intended to represent dynamic running.
%, illustrated in \ref{fig:CTSLIP}. 
The CT-SLIP is an extension of the well-known SLIP model \citep{blickhan1989, movementcriterion, simple, full1999templates}.
We customized the spring force model to be a non-conservative Hill-like muscle model shown in \eqref{eqn:hill-muscle-model}, where the original SLIP had a Hookean spring.The Hill muscle model \cite{Hill} has been postulated to be of sufficient accuracy to be useful for simulating human musculoskeletal behavior \cite{Winters, Bogert}
\begin{equation}\label{eqn:hill-muscle-model}
F(\zeta) = K (L-\zeta)(1+\eta \dot \zeta) - \mu \dot \zeta
\end{equation}

The length $L$ represents the length of the leg at touchdown.
The parameters $K$ and $\eta$ provide averaged approximations to the length dependent and velocity dependent terms (respectively) of the Hill muscle model; $\mu$ adds some dissipation, capturing the overall energy consuming nature of the task.
We elected to employ it over a conservative spring as we want the freedom to inject or remove energy from the system to broaden the set of achievable motions.
\begin{table}[h]
	\caption{Parameters}
	\begin{center}\label{table:ctslip-params}
		\begin{tabular}{|l|l|l|}
			\hline
			Parameter & Definition & Nominal Value  \\ \hline
			$\eta$ &$F_V$ average slope &-0.03 \\ \hline
			$ \mu$ & dissipative loss&0.3\\ \hline 
			$ L$ & $F_L$ average slope &80 \\ \hline
			$t_s$ & torsional spring at hip & 0.1 \\ \hline
		\end{tabular}
	\end{center}
\end{table}

The CT-SLIP has two leg which commutate around the center-of-mass (COM). 
The angle of each leg is determined by a piece-wise monotonic feedfoward (hence, ``clocked'' reference curve that depends only on time - the Buehler clock \citep{simple, saranlirhex2001}. 
There is a torsional spring and actuator at the hip that tracks the reference signal with standard PID control on angle.

The cycling legs, tuned for appropriate parameters, produces a stable forward motion of the COM.
We operationally assume without proof that a phase-like quantity exists for the CT-SLIP, and we use the numerical tool \emph{Phaser} \citep{phaserRevzenGuck2008} (which estimates phase from trajectory data) to produce a workable phase-map from hybrid data. 
%We will present the exact equations of motion the sequel.

In this regime, we aim to recover a stable limit-cycle $\gamma_1$ post-damage that has the same image through the observing function as the original cycle $\gamma_0$.
``Damage'' is modeled as a destructive and irreversible parametric shift that causes the majority of trajectories from uniformly sampled initial conditions to crash (hip-mass striking ground). 

We will restrict our attention to the stance dynamics, as the aerial dynamics are ballistic -- if we recover a desired motion in the stance phases, the aerial phases will be preserved as well as we will only allow in the leg parameters.
%\GC{I might just dump this whole Lagrangian comment -- it's not helpful}.
Using the same dimensionless polar coordinates $\left(\zeta, \psi \right)$ for the configuration space $Q = \Reals \times S^1$, (see Fig. \ref{fig:ctslip-schematic}) as \cite{seipel2007CTSLIP}, the Lagrangian for the CT-SLIP during stance is given by:

%\begin{figure*}
%	\includegraphics[width=\textwidth]{ctslip.png}
%	\caption{The CT-SLIP with indicating variables. For clarity, only one leg is shown. A periodic gait is an concatenation of a sequence trajectory segments from holonomic subsystems. Reproduced from (SEIPEL)}
%	\label{fig:ctslip-schematic}
%\end{figure*}

\begin{equation}\label{eqn:CT-SLIP-Lagrangian-Stance}
L\left(\zeta, \psi, \dot{\zeta}, \dot{\psi},\right) = \frac{1}{2}\dot{\zeta}^2 + \frac{1}{2} \zeta^2 \dot{\psi}^2 - V(\zeta;k)
\end{equation}
The function $V(\zeta; k)$ is the conservative component of \eqref{eqn:hill-muscle-model}.
Since the spring law given in \eqref{eqn:hill-muscle-model} is \emph{not} conservative, our final dynamical system will have a forcing term that accommodates the non-conservative part.
%
%
%\GC{Maybe just write this in terms of the flow to avoid using too much notation to define the hybrid vector field.}
For $z=(x,y,\dot{x}, \dot{y}) \in T Q$, let $\dot{z} = f(z,\lambda, u)$ be the 
resulting Euler-Lagrange equations CT-SLIP hybrid model with forcing function(see  \cite{saranlirhex2001, seipel2007CTSLIP} for full equations of motion) . 
Assume for parameters, described in table \ref{table:ctslip-params}, $\lambda_0 := \left(t_s^0, L^0, \mu^0, \eta^0\right) \in \Pc \subset  \mathbb{R}^4$,  a stable periodic $\Gamma$ exists, with stability basin $\mathcal{B}$. We conflate the geometric object $\Gamma$ with an arbitrary solution on it parameterized by $x_{\gamma}(t)$.

Define the disturbed system by setting $\lambda_1 = \left(t_s^1, L^1, \mu^1, \eta^1\right)$, where $t_s^1 = t_s^0 + \epsilon$. 
We assume that this perturbation is neither small in magnitude, nor reversible.
That is, the damaged system has its torsional spring gain stuck at the value $t_s^1$, and that this value has destabilized $\Gamma$. 

In rectangular coordinates, the desired motion is given by 
 $$\gamma_0(t) := \left(x(t), y(t), \dot{x}(t), \dot{y}(t)\right) \subset T Q, ~t \in \left[0,T\right]$$
For $\lambda_0$, this is a stable limit cycle with phase map $\varphi: \mathcal{B} \to \Image{\gamma_0}=:\Gamma$. 
In this case, $\varphi$ is a asymptotic phase map, where it is defined. 
We are assuming without proof that a phase-like quantity exists away from non-smooth points for the CT-SLIP, and estimate a phase map from trajectory data in simulation using the \emph{Phaser} algorithm \citep{phaserRevzenGuck2008}.
$\varphi$ defines a differential form $\omega$ such that $\omega(f) = 1$ called the temporal 1 form;for details, see \S \ref{appendix:phase-limit-cycle}.

We propose the following design problem.
Let $E \in \mathcal{C}^{\infty}\left( TQ \times \Pc, \Reals\right)$.
Assuming that $z(0) = z_0$, find a  $\lambda_1$ such that the following two equations hold $\forall t \in [0,T]$. 
\begin{subequations}
\label{eqn:ctslip-constraints}
\begin{align}
\frac{d}{dt} \varphi(z_{\lambda_1}) = 1 \label{eqn:ctslip-constraint1}\\
%\omega\left( \dot{z}_{\lambda_1} \right) = 1 \label{eqn:ctslip-constraint1}\\
\frac{d}{dt}E(\varphi(x_{\lambda_1}))  &= \frac{d}{dt} E(x_{\lambda_0}) \iff L_{f(z, \lambda_0)} \left(E \circ \varphi\right) = L_{f(z, \lambda_0)}E \label{eqn:ctslip-constraint2}
\end{align}
\end{subequations}
%%%NOT F A CONSTRAINT IS ON TQ NOT TTQ

The above equations are two equations on the four dimensional state space $TQ$. 
However, we want $(z,v)$ to be an integrable submanifold, i.e., we want $\frac{d}{dt} z = v$.
Including the two additional constraints that $\frac{d}{dt}x=\dot{x}$, and $\frac{d}{dt}y=\dot{y}$, we have a fully determined set of equations on $TQ$, so that a unique curve $z(t)$ obeys them. 
That is, at a point $x \in Q$, there is a unique $v \in T_xQ$ that satisfies Eqn. \eqref{eqn:ctslip-constraints}.
%We are additionally requiring that the $(x,v)$ define a trajectory, rather than just be random points of $TQ$.

So,if $z(0)=\gamma_0(0),~ z(t)=\gamma_0(t)$.
We will omit repeating these last two constraints for the remainder of the section for brevity, as they are straightforward.

Control of the CT-SLIP dynamics is accomplished parametrically - control inputs are restricted to manipulating parameters $\lambda \in \Pc$.
Each execution of the system has a fixed collection of parameters - feedback is not being used to modify the system dynamics as a function of state.
Rather, the control problem is presented as determining fixed values for a collection of parameters whose resultant open-loop dynamics satisfy the afore-mentioned phase and energy conditions. 

It is important to note that $\frac{d}{dt} E(x_{\lambda_0})$ is taking the derivative of $E$ \emph{along the periodic orbit}, while $\frac{d}{dt} E(\varphi(x_{\lambda_1}))$ is projecting a state $x_{\lambda_1}$ that is \emph{off} the orbit to its in-phase companion on the orbit, \emph{then} differentiating. 

The relationship is concisely expressed geometrically. 
Fix $z(t) = z_0$.
Let $r(t) = \frac{d}{dt} E(z_{\lambda_0}) = dE_{\varphi(z)} \left(D \varphi_z f(z,\lambda_0) \right)$.
Let $\omega, dE \in T^*\Gamma$.
Design $\lambda_1 \in \Pc$ such that, for $t \in [0,T]$, 
\begin{subequations}
\begin{align}
\omega_z(f(z,\lambda_1)) = 1 \\
dE_{\varphi(z)} \left(D \varphi_z f(z,\lambda_1) \right)=r(t)
\end{align}
\end{subequations}

The observant reader  will notice this condition as \emph{lifting} the constraints from $\gamma_0$ to the neighborhood where $\varphi$ (and thus $\omega$) is defined.
We are able to equate this condition on differential forms to signals of time by evaluating them along a specific trajectory.
%Also, observe that while the vector field is second order, our control problem is first order, relegating any role forces play to be implicit.
%Equivalently, it is a condition \emph{on solutions}, not \emph{on a vector field}. 
%Ergo, the controller can be designed without explicit knowledge of $f$. 
We will expand on this in the sequel.

\subsubsection{Optimization}
The objective is to fit the Lie derivatives shown in  Eqn. \eqref{eqn:ctslip-constraints} using measured/simulated trajectory data, assuming the that underlying vector field(s) is unknown.
Since we're seeking a parameter  $\lambda_1$ with the irrevocable condition that $t_s = t_s^0+\epsilon$ that satisfies our tracking requirements, we could rephrase the design into a regression to \emph{best fit} the constraints.
I.e, for $t \in [0,T]$, $x_{\lambda}(0) = x_0, ~ \forall \lambda \in \Pc\left.\right|_{t_s=t_s^0+\epsilon}$,
\begin{equation}\label{eqn:ctslip-opt-exact}
  \lambda_1^* = \text{arg} \min_{\Pc\left.\right|_{t_s=t_s^0+\epsilon}} \nonumber \left(\vphantom{\frac{d}{dt}}\norm{\varphi(x_{\lambda}(t))-\varphi(x_{\lambda_0}(t))}^2 \right. +
  \left.\norm{\frac{d}{dt}\left(E \circ \varphi(x_{\lambda}(t))\right)-r(t)}^2 \right)
\end{equation}
The norm $\norm{\cdot}$ in the above the 2 norm on functions.
% i.e.\GC{No we used a modified two-norm}
%\begin{equation*}
%\norm{f} = \sup \left\{ |f(x)| \left.\right| x \in \text{Dom}(f) \right\}
%\end{equation*}

Thus $\lambda_1^*$ is the parameter that minimizes the largest absolute difference over the entire domain of $t$. 
Other function space norms can be selected, but as they are not generally equivalent, the value of $\lambda_1^*$ depends on the choice.

The requirement that $\frac{d}{dt}z=v$ will be automatically enforced by the numerical integrator we will use to generate trajectories with. 

We elect to further modify our constraints. 
If we solve Eqn. \eqref{eqn:ctslip-opt-exact} perfectly, we'd have $x_{\lambda_1}(t) = x_{\lambda_0}(t)$.
We instead relax \eqref{eqn:constraint-fit} to instead be, for some constant $c \in \Reals_{>0}$, 
\begin{equation}\label{eqn:ctslip-relaxed-phase}
\omega_z(f(z,\lambda_1)) = c > 0
\end{equation}

By doing this, the phase rate of $x_{\lambda_1^*}$ is not required to match that of $\gamma_1$, but merely be positive.
Geometrically, this means it must permute the isochrons in the same order, but not necessarily at the same rate.
Physically, this allows the damaged system to potentially have a different frequency than the undamaged system, yet e.g., the sequencing of the limb touchdown sequence is preserved.

\subsubsection{Simulation Results}

The system was simulated in Python 2.7.5 using the NumPy and SciPy open-source numerical libraries.
We take the function $E$ to be the elastic energy stored in the legs. 
Our optimization algorithm of choice is the Nelder-Mead implementation provided by the Scipy \verb|optimize| library.
We are attempting to recover when model information is poor and expensive to determine; Nelder-Mead was elected as it requires no knowledge except function evaluations.
We additionally define the ``total'' energy $E_T$ to be the sum of kinetic and potential energy of the COM.
The functions $E,E_T$ and $\varphi$ were defined on the entire domain by taking a Fourier series approximation of discrete values at sample points produced by numerical integration on the dynamics.
The order of the series was determined by the operator through inspection for quality of fit.

A ensemble of goal trajectories simulated at random initial conditions is generated as a ground truth to match, from which phase and derivatives can be determined without knowledge of the vector field.
The randomly generated initial conditions are fixed at initialization; they do not vary between function calls of the optimizer.
%The fit is computed through optimization. 
The implementation of  \eqref{eqn:ctslip-constraints} subject to the relaxed phase constraint given in Eqn. \eqref{eqn:ctslip-relaxed-phase} was accomplished with the following cost  function.
\begin{align}\label{eqn:ctslip-data-cost}
f(\lambda):=
&\norm{\frac{d}{dt}\left(\frac{E(\varphi(x_{\lambda_1}))}{\langle E_T(x_{\lambda}) \rangle}  - \frac{E(x_{\lambda_0})}{{\langle E_T(x_{\lambda_0})}}\right)}^2
+\alpha \norm{ \frac{\Delta \varphi}{\Delta t}  - \langle \frac{\Delta \varphi}{\Delta t} \rangle}^2 + \beta \norm{\langle \frac{\Delta \varphi}{\Delta t} \rangle }^ {-1}  
\end{align}
$\langle\cdot \rangle$ denotes the mean along a sample path. 
The mean is used to normalize the terms as a proportion of the total average energy so that the same modulation is preserved, rather than  attempting to enforce a particular absolute energy level.

The second term is the variance of the time derivative of phase along sample paths - driving it to zero requires $\varphi$ be constant along solutions.
The last term is the inverse of the mean time derivative of phase - it penalizes phase approaching zero.
In aggregate, the last two terms are attempting for force $\frac{\Delta \varphi}{\Delta t}$ to be a constant bounded away from 0, i.e. - exactly Eqn. \eqref{eqn:ctslip-relaxed-phase}.
$\alpha, \beta \in \Reals$ are  weighting coefficients that we do not argue how to select in a principled manner.
%Fourier series representations of each function are computed to allow same-point comparison in phase.
Similarly, any permutation of the control variables can be formulated analogously.

\begin{figure}
\label{fig:sim-rec}
\centering
\begin{subfigure}{.45\textwidth}
	\includegraphics[width=\linewidth]{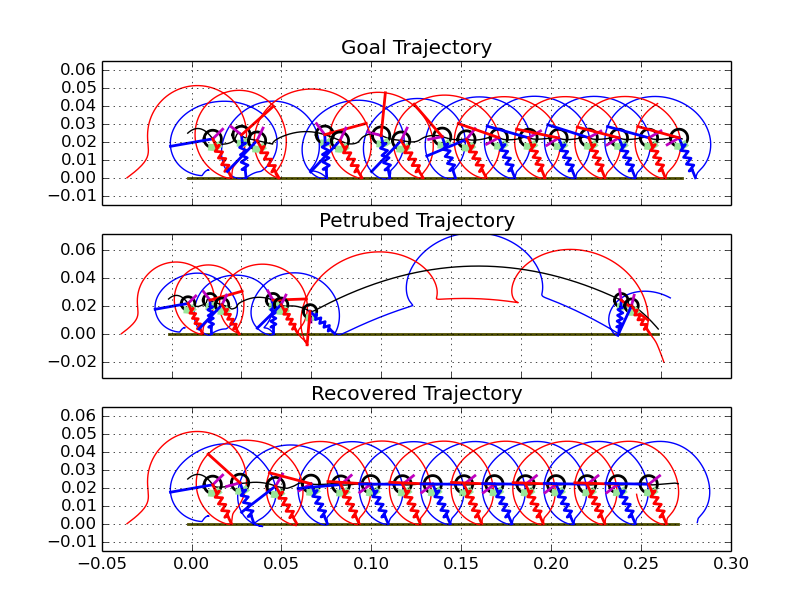} 
	\subcaption{Trajectory 1 of 10} 
	\label{fig:sim-rec-1}
\end{subfigure}
\begin{subfigure}{.45\textwidth}
	\includegraphics[width=\linewidth]{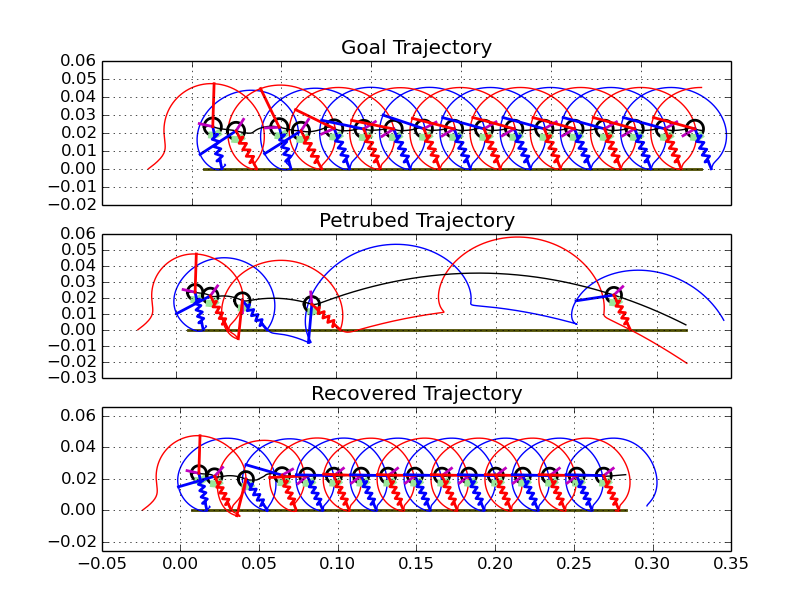} 
	\subcaption{Trajectory 3 of 10} 
	\label{fig:sim-rec-2}
\end{subfigure}
\caption{Traces of the COM and legs. The COM is black, one leg is red, and the other is blue. Ten initial conditions in the stability basin of the original periodic orbit were sampled. The recovery method attempted to find a single $\lambda_1^*$ such that all were stable. }
\end{figure}

Shown in figures \ref{fig:sim-rec-1}, \ref{fig:sim-rec-2} are integration results for two initial conditions (out of 10) that were used to generate the fitting ensemble. 
For the goal system, all ten initial conditions stabilized to a periodic solution. 
For the perturbed system, seven out of ten initial conditions lead to a crash (wherein the COM impacts the ground, and the simulation ceased).
For the recovered solutions, nine out of ten initial conditions recovered to a periodic solution. 

%\begin{figure}[ht] \label{fig:sim-rec-3}
%	\includegraphics[width=0.5\linewidth]{fwith_phase_3.png} 
%	\caption{Trajectories 7 of 10} 
%\end{figure}

We see that the recovered system \emph{on average} has superior performance over the perturbed system, but that the recovered system still does not match the unperturbed system's performance.
The reason for the failure to fully recover is, as of yet, unknown, and some discussion will be given in the sequel to this matter.

\subsection{Theory}\label{sec:theory}

We have seen in \S \ref{sec:ctslip} that the constraint equations in Eqn. \eqref{eqn:ctslip-constraints} provide a complete description of a desired limit cycle.
By enforcing them, a specified behavior was engendered from the CT-SLIP.
However, there are certainly unanswered questions.
Does asymptotic phase help in some special way? What properties do the constraints need to satisfy? What if the control isn't parametric?

We now present a class of mechanical systems relevant to locomotion that provably have a similar property, and furthermore, the constraints will always be on $TQ$, rather than on $TTQ$.
In order words, we will show that there exist Pfaffian-like constraints that we can impose on position and velocity to restore a behavior.
A significant feature of doing so is that the trajectories of mechanical systems live on $TQ$.
By expressing our objective on $TQ$, it is independent of any underlying vector field.
In other words, our approach generates a control objective that is agnostic to the specific dynamics that govern the robot.

\subsubsection{Geometric Mechanics}\label{sec:geo-mech-review}
We now present a brief overview of geometric mechanics an its relevance to locomotion.
A considerable body of literature exists on this topic - the interested reader should consult \cite{bloch1996nonholonomic,ostrowski1996mechanics,bloch2003nonholonomic} and the references therein for precise details. 

Assume that we have configuration space $Q=S \times G$, where $G$ is a Lie group that left acts on $Q$ freely and properly via $\Phi_g$, along with independent Pfaffian constraints $\omega(q) \cdot \dot{q} = 0 \in \mathbb{R}^k$, and Lagrangian $L : \T Q \to \mathbb{R}$. 
We critically assume that both $L$ and the constraints $\omega$ are \emph{symmetric under} $G$, i.e. 
\begin{equation}\label{eqn:sym-lagrangian}
L(q, \dot{q}) = L (\phi_g(q), D_q \phi_g \dot{q})
\end{equation}
\begin{equation}\label{eqn:sym-constraints}
\omega(q) \cdot \dot{q} = D_q \phi_g^T \cdot \omega_{\phi_g(q)} \dot{q}
\end{equation}

If $k = \dim(S)$, in can be shown 
\citep{ostrowski1998geometric, bloch2003nonholonomic,bloch1996nonholonomic,koon1997geometric,hatton2011introduction}. 
that there exists a map $A : TS \to \mathfrak{g}$ such that 
\begin{equation}\label{eqn:reconstruction}
g^{-1} \dot{g} = -A(s) \cdot \dot{s}
\end{equation}

Thus, the body velocity $g^{-1} \dot{g}$ \footnote{by which we mean $g^{-1}\dot{g} : = \xi = T_{g} L_{g^{-1}}\dot{g}$, which is the velocity of the body frame in body coordinates.}  is a algebraic function of shape $s$ and shape velocity $\dot{s}$. 
Physically, $k = \dim(S)$ describes a kinematic system - the motion of the robot is completely determined via shape, with no drift due to momentum.

We now exclusively consider curves $s(t)$ that are periodic with period $T$. 
Eqn. \eqref{eqn:reconstruction} is a differential equation on the group $G$ that defines the \emph{holonomy} or \emph{horizontal lift} \citep{kobayashi1963foundations1} of a path $s(t), t \in [0,T] \in S$. 
While the dynamics of $s$ are unimportant to validity of this representation, we will always assume that the dynamics of $s$ are controlled, i.e. there exists some equation 
\begin{equation*}
\dot{s} = f(s) + G(s)u
\end{equation*}

That is, the shape variables are a first-order control-affine system that evolves dynamically, but there are sufficient constraints that there is a algebraic relationship between the body velocity and shape state.
\footnote{Some authors take it a step further, and simply have $\ddot{s} = u$.}
A solution curve $c(t) := (s(t),g(t))$ is called \emph{horizontal} by virtue of satisfying $\omega(c(t)) \cdot \dot{c}(t) = 0$. 
The curve $c(t)$ is a path connecting $g(0)$ to $g(T)$ that satisfies the constraints, with holonomy $g(T)g(0)^{-1}$. 
The path $c$ is not necessarily unique - the set of all curves that achieve a desired group motion is a Lie group in its own right \citep{kobayashi1963foundations1}.
\begin{equation}
\text{Hol}(m) = \{g_s \left.\right| s:[0,T] \to S, s(0) = s(1)  \}
\end{equation}

Solutions are generally difficult to determine, and can be expressed by various expansions, such as a path-ordered exponential \cite{marsden1990reduction} or Magnus expansion, \cite{radford1998local}

In the case that the group $G$ is abelian, conventional integration suffices \cite{marsden1990reduction}
\begin{equation*}
g(t) = \exp\left(\int_0^t \xi(s) ds\right)
\end{equation*}

Classic examples would include the kinematic car \citep{ pepy2006path} or Purcell swimmer \citep{hatton2013geometric}; a classic counter-example is the snakeboard \citep{bloch1996nonholonomic}.

The group motion $g$ is the displacement in the world, while the $s$ (shape) variables describe relative motion of the robot limbs with respect to the center of mass. 
For a closed loop $s(t), t \in [0,T]$, a single revolution produces an holonomy element $h:=g(T)g(0)^{-1}  \in G$.
In mechanics, this is refereed to as the  \emph{geometric phase} \citep{marsden1990reduction} of the loop $s(t)$.
Repeated iterations in $S$ simply concatenate the group elements $h$ owing to its group structure.
In this sense, locomotion is well represented - repeated cycling of limb motions produces displacement in the world, agnostic to starting position. 
In general, the same group motion can be effected by many  gaits.
\subsubsection{Implicit Functions}
To motivate the construction we present in the sequel, consider the following straightforward problem. 
Let $N$ be a manifold of dimension $n$, and assume there exists flow $\phi^t : N \to N$, along with $C^1$ function $F : N \to \mathbb{R}^k$.
Suppose that we have initial condition $y(0) \in N$ such that $F(\phi^t(0) )= 0 \in \mathbb{R}^k, \forall t \in [a,b]$.  
In a local coordinates, we have $\phi^t(y(0)) =: \left(y_1(t),\dots, y_n(t)\right)$ , $F(y_1(t),\dots, y_n (t))=0$.
Assume that $F$ is full rank at each point $y(t)$, so that by the implicit function theorem, there exists $C^1$ function $h$ such that
\begin{equation*}
F(h(y_{k+1},\dots,y_n), y_{k+1},\dots y_n)=0
\end{equation*}

Assume now that there exists submersion $\varphi : M \to N$.
We may then consider the pullback $\varphi^*h(m) := h(\varphi(m)_{k+1}, \dots, \varphi(m)_{n})$.
Suppose we found a curve $m(t) \subset M$ such that
\begin{equation}
\varphi^*h(m(t)) = \left(y(t)_{k+1},\cdots,y(t)_{n}\right)
\end{equation}
and 
\begin{equation}
\left(\varphi(m(t))_{k+1},\dots,\varphi(m(t))_{n}\right) = \left(y(t)_{k+1},\cdots,y(t)_{n}\right)
\end{equation}
Then,
\begin{equation}
F(\varphi(m(t)) = 0
\end{equation}

While the above is completely trivial, it expresses the design problem of finding $m(t)$ as two testable objectives -  insist that $\varphi(m(t))$, the phase of $m(t)$, is a specified reference, and that $m(t)$ satisfies the constraint defined $\varphi^*h$. 
As neither the kernel of $\varphi$ or $\varphi^*h$ is non-trivial, there are many possible equivalent candidates for $m(t)$. 
If an initial solution curve $m_1(t) \subset M$ is no longer achievable due to damage, we can resolve the problem in for another curve $m_2(t)$ that \emph{exactly} solves  $F = 0$.
We now apply this idea to our problem in geometric mechanics.

\subsubsection{The Kinematic Case}\label{sec:kin-case}
Following our approach in \ref{sec:geo-mech-review},
let $Q = S \times G$ for lie group $G$, with $G$-invariant Lagrangian $L : \T Q \to \Reals$.
Everything will be assumed to be $C^{\infty}$ unless otherwise stated.
Assume that have Pfaffian constraint form $\omega \in \Omega^k$, i.e, a collection of $k$ linearly independent 1-forms $\omega_i, i = 1,\dots,k$ such that $\omega_i(q) \dot{q} = 0$ for all  solutions  $(q,\dot{q})$ determined by $L$. 
We have that $\omega$ is $G$-equivariant. 
In coordinates, this implies we may write it as, for $\dot{q} = (\dot{r}, \xi=g^{-1}\dot{g})$ :
\begin{equation*}
\omega_{\xi}(r) \xi + \omega_{\dot{r}}(r) \dot{r} = 0 \in \Reals^k
\end{equation*}
We assume that $k = \dim(G)$, so that we are in the principal kinematic case. 
Assume $\dim(S) = n > k$, defining gauge field $A$ given by, for $\omega_{\xi} \in GL(k,k), ~\omega_{\dot{r}} \in L(n,k)$.
\begin{equation*}
\xi = -\omega_{\xi}^{-1} \omega_{\dot{r}} \cdot \dot{r} =: -A(r) \cdot \dot{r}
\end{equation*}
\subsubsection{Encoding Templates and Recovery}
Let $\hat{Q} = \hat{S} \times G$, and assume that we have map $\varphi:\hat{Q} \to Q$ such that $\varphi(\hat{s},g) = (\varphi^s(\hat{s}),g)$, i.e. it does not depend on $G$. 
We then have a map $\varphi^s : \hat{S} \to S$.
Given the constraint form $\omega:TS \times \mathfrak{g} \to \mathbb{R}^k$,we may pull back to $\varphi^* \omega(\hat{s}) \cdot (\dot{\hat{s}},\xi) = \omega(\varphi^s(\hat{s}))  D\varphi^s \cdot  (\dot{\hat{s}}, \xi)$.
We use $\xi := g^{-1}\dot{g}$ for the body velocity.

We assume that we are given curve $q(t)=(s(t),g(t)) \subset Q$ such that $\omega(q) \cdot \dot{q} = 0$.

Suppose that we had curve $\hat{q}(t)=(\hat{s}(t),\tilde{g}(t)) \subset \hat{Q}$ such that
\begin{equation}\label{eqn:phase-fit}
\varphi^s(\hat{s}) =:v(t) =  s(t)
\end{equation}
and that 
\begin{equation}\label{eqn:constraint-fit}
\varphi^*\omega(\hat{s}) \cdot (\dot{\hat{s}},\tilde{g}^{-1}\dot{\tilde{g}}) = 0.
\end{equation}

Eqn. \eqref{eqn:constraint-fit}, at each point $\hat{s}$, can be though of as $\dim{G}$ linear equations in $\dim{Q} \times \dim{G}$ unknowns.
By assuming Eqn. \eqref{eqn:phase-fit} holds, we also have
\begin{equation*}
\frac{d}{dt} v(t) = \frac{d}{dt} s(t)
\end{equation*}

Ergo,along $v(t)$ and $s(t)$ , $\omega(v(t),\cdot) = \omega(s(t),\cdot)$. 
Thus, we have $\dim{G}$-equations in $\dim{G}$ unknowns. 
We initially assumed the problem had rank $\dim{G}$, so that there is a unique solution, so, for $\tilde{\xi} = \tilde{g}^{-1}\dot{\tilde{g}}$, 
\begin{equation}\label{eqn:recovery-rank-xi}
\omega(s,\xi) = 0=\omega(v,\tilde{\xi}) \implies \xi = \tilde{\xi}.
\end{equation}

We call the codomain $Q$ for phase map $\varphi$ the \emph{encoding template}.
It represents a simpler mechanical system (as we assume $\dim{Q} < \dim{\hat{Q}})$ that is still capable of executing the desired $\xi(t)$. 

An important feature of the system on $Q$ is that we \emph{only} require the above relationships for $\xi(t)$. 
The template system may be able to engage in behaviors that the anchor system is \emph{not} able to do. 
Not all trajectories can or need to be lifted to trajectories of the anchor system. 

\subsubsection{Missing Constraints}\label{sec:missing-constraints}
Assume that we have a known embedded curve $(r_0(t),g_0(t)) \subset S \times G,~ t \in (t_a, t_b)$ such that $\xi_0(t) :=g_0^{-1}(t)\dot{g}_0(t) = A(r_0(t)) \cdot \dot{r}_0(t).$
Since we have a ``wide'' matrix $A$, there are infinitely many curves $r(t)$ that map to $\xi_0(t)$, as by fiat we assume there exists at least one $r_0$.
For $\Gamma := \text{Im}(r_0(t),g_0(t))$, assume that $\rank{A\left.\right|_{\pi_1 \circ \Gamma}} = s$ is constant. 
Since rank is an open condition, there exist tubular nhbd $U \subset S$ of $r_0(t)$ s.t. $\rank{A\left.\right|_{U}} = s$ is constant.

Let $f(r,g) : S \times G \to \Reals$ be a real-valued function, then 
\begin{equation*}
f(r_0(t),g_0(t)) = \eta(t)
\end{equation*}
have retraction $\varphi : U \to \Gamma$, and define the pullback 
\begin{equation}
\tilde{f}(r,g) = f \circ \varphi
\end{equation}
so that by definition,
\begin{equation}
\tilde{f}\left.\right|_{\Gamma}=f
\end{equation}
Implicit in the definition of $\varphi$ is the time-parameterization of $(r_0,\xi_0)$.
Explicitly, $\theta(t) = (r_0(t),g_0(t))$ is the embedding of $(t_a,t_b)$
\begin{equation}
\tilde{f}(r,g) = f(\pi_1 \circ \varphi(r,g), \pi_2 \circ \varphi(r,g)) = \eta\left(\theta^{-1} \circ\varphi(r,g)\right)
\end{equation}
We write via chain rule
\begin{align}
d\tilde{f} \cdot \left(\dot{r}_0(t), \dot{g}_0(t) \right) &= df \cdot \left(\dot{r}_0(t), \dot{g}_0(t) \right)\\ &=df \cdot \left(\dot{r}_0(\theta^{-1}\left(\left(r_0,g_0\right)\right), \dot{g}_0(\theta^{-1}\left(\left(r_0,g_0\right)\right)  \right)\\ &=:  \gamma \left(r_0,g_0\right)
\end{align}

Since $\tilde{f}$ is defined everywhere on $U$, so is its derivative $d\tilde{f}$.
This allows us to formulate the constraint 
\begin{equation}
d\tilde{f}\cdot\left(\dot{r}, \dot{g}\right)= \gamma(\varphi(r,g))
\end{equation}

Assume now that the function $\varphi(r,g)$ is group-invariant, and that $f$ generates a foliation of $Q$ that is permuted by $\phi_h$, so that ,
\begin{equation*}
\tilde{f}(r,g) = \tilde{f}(r, \phi_h(g))+c(h)
\end{equation*}
Where $c : G \to \mathbb{R}$ is some function as smooth as $f$. 
Differentially, this implies $\forall r \in S, g,h\in G$, where $d\tilde{f}=\partial_r \tilde{f}(r,g) dr  + \partial_g \tilde{f} dg$
\begin{align}
&\partial_r \tilde{f} (r,g) dr = \partial_r(r,\phi_h g) dr \\
&\partial_g \tilde{f} (r,g) dg = \partial_g \tilde{f} (r,\phi_h g) dg \cdot  d\phi_h   
\end{align}unique

The first equation implies $\partial_r \tilde{f}$ does not depend on $g$.
The second equation implies, for $h=g^{-1}$, 
\begin{equation*}
\partial_g \tilde{f}(r,g) = \partial_g\tilde{f}(r,e) d\phi_{g^{-1}} = \partial_g \tilde{f}(r,e)\cdot g^{-1}
\end{equation*}
where in the last statement we have conflated the action and lifted action of $G$.

These together let us write 
\begin{equation}\label{eqn:newform}
\partial_r \tilde{f} (r) dr + \partial_g \tilde{f}(r,e) dg \cdot g^{-1} = \gamma(r)
\end{equation}

We have $\gamma(r)$ as by assumption $\varphi(r,hg) = \varphi(r,g)$, implying it is only a function of $r$.
We simplify notation and write \eqref{eqn:newform} as, for $\xi = g^{-1}\dot{g}$, and implicitly assuming that it is along curve $(r(t), g(t))$
\begin{equation}\label{eqn:function-constraints}
\tilde{\omega}_r(r) \dot{r} + \tilde{\omega}_{\xi}(r) \xi = \gamma(r)
\end{equation}

Key assumption: We now assume that $\exists i \in \{1,\dots,k\}$ such that the known constraint $\omega_i$ is no longer in effect - i.e. - solutions curves are not required to satisfy it. 

Since the 1-form determined by \eqref{eqn:newform} is group-equivariant, we can consider the quantity, $i = 1,\dots,k-1$
\begin{align}
\omega_{1\xi}\xi + \omega_{1\dot{r}} \dot{r} &= 0 \\ 
\vdotswithin{=} \notag \\
\omega_{(k-1)\xi}\xi + \omega_{(k-1)\dot{r}} \dot{r} &= 0 \\
\tilde{\omega}_r(r) + \tilde{\omega}_{\xi} \xi &= \gamma(r) \label{eqn:augment-cons}
\end{align}

where we have replace $\omega_i$ with $d \tilde{f} = \gamma(r)$. 
Call this form $\hat{\omega}$ - we have generated a form by pulling back $f$ into $M$.
Observe $\hat{\omega}$ is \emph{still kinematic} - i.e. - by obeying it, the group element $\xi$ is enforced. 
However, for the implication in Eqn. \eqref{eqn:recovery-rank-xi} to hold, the rank of  \eqref{eqn:function-constraints} needs to be $\dim{G}$. 

\subsubsection{Do we require kinematic systems?}

An \textbf{key} aspect of \S \ref{sec:missing-constraints} is that we no longer require $\omega(q) \cdot \dot{q} = 0$, but rather we allow \emph{affine} constraints. 
In fact, we never needed to assume Pfaffian constraints as in \ref{sec:kin-case}, and could've just started with affine constraints.
However, we elected to begin with kinematic systems due to their relative familiarity.

In this sense, the reconstruction equation approach only requires that the dynamics are \emph{fully} constrained in the group directions, rather than \emph{kinematic} in the sense that they have no history -- the ``shape'' variables could have arbitrary dynamics, but if there are enough constraints, there is still an \emph{algebraic} relationship to the group motion.

Ergo, our approach works for any constrained mechanical system that is invariant under the action of a group, rather than those that are properly \emph{kinematic}.

\subsubsection{Transversality of Group Invariant Functions}
We now show that completing the constraints with differentials such that the rank is expanded is a generic property.
We use $L^r(n,k)$ to denote the $n \times k $ matrices that have rank $r$, and $L(n,k) = \bigcup_{r}L^r(n,k)$
Consider an $k\times n$ matrix-valued function $A:S\to \text{Hom}(V,W)$
\begin{equation}A=
\begin{bmatrix}
a_1(s) \\
a_2(s) \\
\vdots\\
a_{k}(s)
\end{bmatrix}
\end{equation}
Each $a_i, i \in 1,\dots,k$ denotes a row. 
Assume that $A(s)$ is of rank $k$ for all $s\in S$.
Let $f \in C^{\infty}\left(S,\text{Hom}\left(\mathbb{R}^n,\mathbb{R}^N\right)\right)$ be a smooth function, and define
\begin{equation}
B(s):=
\begin{bmatrix}
A(s) \\
f(s)
\end{bmatrix}
\end{equation}
I.e. we augment the rows of $A$ with a $N$ new rows determined by the function $f$.
We would like to determine when our choice of $f$ will have the rank of $B \geq k+1$.

We make use of the following theorem.

\begin{theorem}[Prop 5.3, \citep{golubitsky2012stable}] $L^r(V,W)$ is a closed submanifold of $\text{Hom}(V,W)$ with $\text{codmin}(L^r(V,W))=(m-q+r)(n-q+r)$ where $q = \min(m,n)$.
\end{theorem}
The dimension of $B$ is $(k+N)\times n$.
We have $m = k+N, n = n, q=m, r=k+N-k=N$.
Since adding a row cannot lower the rank of a matrix, we equivalently want to ensure that $B \notin L^N$.
So, $\text{codim}(L^N) = N(n-k)$. 
We now state a relatively apparent fact about transverse maps.
\begin{theorem}[Prop 4.2, \citep{golubitsky2012stable}]
	Let $X$ and $Y$ be smooth manifolds, $W \subset Y$ a submanifold. 
	Suppose $\dim{X} + \dim{W} \leq \dim{Y}.$ (i.e. - $\dim{X} \leq \text{codim}(W)$)
	Let $f : X\to Y$ be smooth and suppose that $f$ is transverse to $W$. 
	Then $f(X) \cap W = \varnothing.$
\end{theorem}

So by the prop, a sufficient condition  for $B(s) \notin L^N$ for all $s \in S$ is: Assume that $B$ is transverse to $L^N$, and  $N(n-k) > n$
If so, the only way for $B$ to successfully transverse to $L^N$ to have $B(s) \notin L^N$ by definition, as $\dim{\text{Im}(B)} = n$, $\dim{L^N} = N+k - N(n-k)$.
Ergo, if $B$ is transverse
\begin{equation}\label{eqn:transverse-N}
N > \frac{n}{n-k}
\end{equation}
Then $B(s)$ will have rank greater than $k$.

We are bailed out by the following:

\begin{theorem}[Cor. 4.12, \citep{golubitsky2012stable}] Let $X$ and $Y$ be smooth manifolds, and let $W$ be a submanifold of $Y$. 
	Then, the set of smooth mappings of $X$ to $Y$ which intersect $W$ transversely is dense in $C^{\infty}(X,Y)$, and if $W$ is closed, then this set is also open.
\end{theorem}

$L^N$ is closed, so by the above there is an open and dense set of maps $f$ such that $B$ will be rank greater than $k$.

We now modify the problem to ask a slightly different question.
Suppose instead picking functions that directly provide rows of $B$, we wanted to pick $N$ functions $f \in C^{\infty}(S,\mathbb{R})$ such that the matrix-function extension,
\begin{equation}
B(s):=
\begin{bmatrix}
A(s) \\
df(s)
\end{bmatrix}
\end{equation}
had $\rank{B(s)} \geq k$?

Let $\sigma \in J^1(X,Y)$; let $f$ represent $\sigma$.
$df_p$ is a linear mapping. Define $\rank{\sigma} = \rank{df_p}$.
Define the set 
\begin{equation*}
S_r := \{ \sigma \in J^1(X,Y) \left.\right| \text{corank}(\sigma) = r \}
\end{equation*}
Then we have the following theorem.
\begin{theorem}[Prop 5.4, \citep{golubitsky2012stable}]$S_r$ is a submanifold of $J^1(X,Y)$ with $\text{codim}(S_r) = (n-q+r)(m-q+r).$
	In fact, $S_r$ is a sub bundle of $J^1(X,Y)$, with fiber $L^r(\mathbb{R}^n,\mathbb{R}^m)$
\end{theorem}

If we take $W  = S^r$, we want $\sigma$ transverse to $W$.\
Since $J^1(X,Y)$ is a manifold, and $S^r$ is closed, we immediately see by the previous that the space of jets that are transverse to $W$ is open and dense.
Motivated by this observation, we  select $N > \frac{n}{n-k}$ differentiable functions to have rank greater than or equal to $k+1$.

We have shown that we can add rank to a set of constraints by using the derivative of some $f \in \mathcal{C}^{\infty}(S, \mathbb{R}^N)$. 
However, for our problem, we want the $f$ we use to have \emph{group-invariant} constraints, like above, so that we still have a connection. 

We have that if the foliation of the level sets of $f$ are permuted by $\phi_h$,
\begin{equation}
f(r,\phi_h \cdot g) - f(r,g) = c(h)
\end{equation}
Let $g = e$, so that 
\begin{equation}
f(r,h) = c(h)+f(r,e)
\end{equation}
$c \in \mathcal{C}^{\infty}(G,\mathbb{R})$, and since $e$ is fixed, we think of $f_e (r) \in f(R,\mathbb{R}^n)$ as a pure function of shape.
Ergo, let $\hat{c} \in  \mathcal{C}^{\infty}(G,\mathbb{R}$, $\hat{f} \in f(S,\mathbb{R}^n)$.
We can define a function by
\begin{equation}
f(r,g) := \hat{c}(g) + \hat{f}(r)
\end{equation}
Then,
To see that $f(r,\phi_h \cdot g) - f(r,g) = c(h)$, we compute
\begin{equation}
f(r,\phi_h \cdot g) = \hat{c}(\phi_h \cdot g) + \hat{f}(r)
\end{equation}
\begin{equation}
f(r,\phi_h \cdot g) -  f(r,g) = \hat{c}(\phi_h \cdot g) - \hat{c}(g) = \hat{d}(h)
\end{equation}
Where $\hat{d}$ is a function only of $h$.
Thus, let $c_{f} \in \mathcal{C}(G \times S, \mathbb{R})$, $c_f(g,s) = c(g) + f(s)$.

Since there is an open and dense set of functions $f$ that provide the necessary rank condition, and we see that we can ``complete'' $f$ to $c_f$ by picking $c$, we see that there is an open and dense set of group invariant functions that meet our rank requirement.

\subsubsection{Energy}

For mechanical systems, energy is a natural choice of constraints, thus we use the following results from \citep{abraham2008foundations}

\begin{definition}[Def. 3.5.5]
	Let $\omega_0$ be the cananonical sympletic form on $T^*Q$. For Lagrangian $L$, let the Lagrangian two form be defined as $$ \omega_L : = (FL)^* \omega_0$$.
	Where $FL :TQ \to T^*Q$ is the fiber derivative of $L$. 
\end{definition}

\begin{theorem}[Def. 3.5.11]
	Given $L : TQ \to \mathbb{R}$, define the \emph{action} $A : TQ \to \mathbb{R}$ by $A(v_x) = FL(v_x) \cdot v_x$ and the \emph{energy} by $A - L$. By a \emph{Lagrangian Vector Field for L} we mean a vector field $X_e$ on $TQ$ such that $i_{X_e}  \omega_L = dE$, for $\omega_L$ in the Lagrangian two form.
\end{theorem}

Suppose then, that we had two lagragians, $L$, and $\tilde{L}$, with associated vector fields $X_e$ and $\tilde{X}_e$.
We'll further assume that $\tilde{L}$ has tunable parameters $p$, so that $\tilde{L}(p)$ yields $\tilde{X}_e(p)$ -- we will imagine these parameters being quantities such as mass, spring stiffness, and the like.

%We can then consider $$i_{X_e} i_{X_e} \omega_L = i_{X_e} dE = dE (X_e)$$.

Pick a point $x \in TQ$, and consider a subspace $A \subset T_xTQ$ of dimension $n < \dim{TQ}$, with basis $\{e_1, \dots, e_n\}$ that spans $A$.
%We assume that $dE \left.\right|_{A} = d\tilde{E} \left.\right|_A$.
%This is merely the assumption that the systems $X_e$ and $X_{\tilde{e}}$ have the same template definition.

We seek a value of $p$, say, $p^*$, then, so that, expanding $X_e$ and $X_{\tilde{e}}$ on $A$, i.e., for linear projection $P : T_xQ \to A$, writing $P \circ X_e$ and $P \circ X_{\tilde{e}}$ in coordinates as:
\begin{align*}
  P \circ X_e = x_1 e_1 + \cdots + x_n e_n \\
  P \circ X_{\tilde{e}} = \tilde{x}_1 e_1 + \cdots + \tilde{x}_n e_n 
\end{align*}
we get problem:
\begin{equation*}
   dE( x_1 e_1 + \cdots + x_n e_n) =  d\tilde{E}(\tilde{x}_1 e_1 + \cdots + \tilde{x}_n e_n )(p).
\end{equation*}

Such a problem has infinite solutions if it has any. 
We strengthen the requirements to:
\begin{equation*}
  \forall i, ~ x_i dE(e_i) = \tilde{x}_id\tilde{E}_i(e_i)(p)
\end{equation*}

Since our ultimate goal is to deduce $P \circ X_e = P \circ X_{\tilde{e}}$, we still run afoul that both $\tilde{E}$ and $X_{\tilde{e}}$ \emph{both vary with} $p$.
Herein we make use of our first assumption -- that we have an encoding template that lifts into $A$.
Equivalently, that
$$dE \left.\right|_{A} = d\tilde{E} \left.\right|_{A}$$
Then, only $X_{\tilde{e}}$ depends on $p$, and the desired result follows. 
\subsection{Crawler Methods}\label{sec:crawler}

As given in \S \mref{sec:crawler} of the main text, the crawler is subject the rigid constraints:
\begin{align}
l_1 = \sum_{i=1}^3 e^{i\sum_{j=1}^k \theta_j}=:f_1(\theta)\label{eqn:crawler-anchor-constraints1}\\
l_2 = \sum_{k=4}^6 e^{i\sum_{j=1}^k \theta_j}=:f_2(\theta)\label{eqn:crawler-anchor-constraints2}
\end{align}
I.e., the robot is subject to the constraints $f_1(\theta)-l_1 =0$, and $f_2(\theta)-l_2 = 0$, where functions $f_i : T^6 \to \mathbb{C}$ are the R.H.S of Eqns. \eqref{eqn:crawler-anchor-constraints1}, \eqref{eqn:crawler-anchor-constraints2}.
%The choices of parameters $l_1,l_2,h_1$, and $h_2$ are shown in Table. \ref{table:crawlers-params}.
%\begin{table}[h]
%	\caption{Parameters}
%	\begin{center}\label{table:crawlers-params}
%		\begin{tabular}{|l|l|}
%			\hline
%			Parameter & Value  \\ \hline
%			$l_1$ & $+2.5+i2$ \\ \hline
%			$l_2$ & $-2.5+i2$\\ \hline 
%			$h_1$ & $ +1 $\\ \hline
%			$h_2$ & $-1$ \\ \hline
%		\end{tabular}
%	\end{center}
%\end{table}

Define the point $q \in \mathbb{C}$ be the midpoint of the two feet, i.e.  $$q = \frac{l_1+l_2}{2}$$

We take $(r,\alpha) \in \Reals \times S^1$, and define our template system with this point $q$ via Eqn. \eqref{eqn:crawler-template-basis}.
\begin{equation}\label{eqn:crawler-template-basis}
r\exp(i(\theta_0+\alpha))+x+iy = q
\end{equation}

Eqn. \eqref{eqn:crawler-template-basis} defines an implicit relation between the template shape variables $(r,\alpha)$ and the anchor shape variables $\theta$. 

Since the template system is the distance $r$ of the COM to the point $q$, while $\alpha$ is the angle of the COM with respect to the center of mass.
%See \ref{fig:crawler-template} for a visualization. \GC{I can probably just modify crawler drawing to have this}

Define the two auxilary functions $p_1 : T^6 \to \mathbb{C}$, and $p_2 : T^6 \to \mathbb{C}$  as:
\begin{align*}
p_1(\theta) :=h_1+\exp(i\theta_1)(1+\exp(i\theta_2)(1+\exp(i\theta_3)))\\
p_2(\theta):=h_2+\exp(i\theta_4)(1+\exp(i\theta_5)(1+\exp(i\theta_6)))
\end{align*}
By solving \eqref{eqn:crawler-template-basis} for $(r,\alpha)$ in terms of $\theta$, 
it can be shown that the phase map $\varphi^s : T^6 \to (r,\alpha) \in \Reals \times S^1$ is given Eqn. \eqref{eqn:crawler-sphase}.
\begin{equation}\label{eqn:crawler-sphase}
\varphi^s(\theta)=\left(\norm{\frac{\left(p_1+p_2\right)}{2}},\phase{\frac{\left(p_1+p_2\right)}{2}}\right)
\end{equation}

The equations  \eqref{eqn:crawler-anchor-constraints1}, and  \eqref{eqn:crawler-anchor-constraints2} are equivalently holonomic constraints on the configuration space -- they induce four real-valued constraint equations on the tangent bundle by differentiation. I.e., 
\begin{equation}\label{eqn:crawler-anchor-pfaff}
\begin{bmatrix}
\nabla_{(g,\theta)}(\text{real}(f_1))\\
\nabla_{(g,\theta)}(\text{imag}(f_1))\\
\nabla_{(g,\theta)}(\text{real}(f_2))\\
\nabla_{(g,\theta)}(\text{imag}(f_2))\\
\end{bmatrix}
\begin{bmatrix}
\dot{g} \\ \dot{\theta}
\end{bmatrix}
=
\begin{bmatrix}
0 \\ 0 \\ 0 \\ 0 
\end{bmatrix} 
\end{equation}
We write \eqref{eqn:crawler-anchor-pfaff} compactly as   
\begin{equation}
\omega_A(g,\theta) \cdot (\dot{g},\dot{\theta}) = 0 \in \Reals^4.
\end{equation}
$\omega_A \in \left(T^*(T^6 \times SE(2))\right)^4\cong \Reals^{4\times9}$ is the Jacobian matrix which multiplies  $(\dot{g},\dot{\theta}) \in T(T^6 \times SE(2))\cong\Reals^9$. 
In the $(g,\theta)$ coordinates, the jacobian appears in block form as:
\begin{equation}\label{eqn:crawler-block-cons}
\begin{bmatrix}
\omega_A^g & \omega_A^{\theta} 
\end{bmatrix}
\begin{bmatrix}
\dot{g} \\ \dot{\theta}
\end{bmatrix}
=
0 \in \Reals^4, ~~\omega_A^g \in \mathbb{R}^{4 \times 3}, \omega_A^{\theta} \in \mathbb{R}^{4\times 6}
\end{equation}

Similarly, the $\varphi(g, \varphi^s(\theta))=(g, r, \alpha)$ template coordinates have two Pfaffin constraints determined via Eqn. \eqref{eqn:crawler-template-basis}, which we write as.
\begin{equation}\label{eqn:crawler-template-cons}
\omega_T(g,r, \alpha) \cdot 
\left(\dot{g} , \dot{r} , \dot{\alpha} \right) = 0 \in \Reals^2
\end{equation}
We denote the associated block structure via $\omega_T^g \in \Reals^{2\times3}$, $\omega_T^s \in \Reals^{2\times2}$, analogous to Eqn. \eqref{eqn:crawler-block-cons}.

If the block $\omega_T^g$ were invertible, we could write a world-coordinate version of the reconstruction equation shown in  Eqn. \eqref{eqn:reconstruction} as $\dot{g} = -\left(\omega_T^g\right)^{-1} \omega_T^{\theta}$.
We note that we have $\dot{g} \in T_g SE(2)$, as opposed to $\xi \in \mathfrak{se}(2)$ as literally shown in Eqn. \eqref{eqn:reconstruction}.
While we could transform $\dot{g}$ to the lie algebra via $T_{g} L_{g^{-1}}$, our expression is sufficient for our example as-is. 
We are only considering once ``stance-phase'' where world coordinates are adequate.
\footnote{Furthermore, it illustrates that our approach of virtual constraints on template models is more general that the kinematic lifting shown in \S \ref{sec:missing-constraints}. 
The construction presented there is \emph{sufficient} and \emph{explicit}, but not the only case where virtual constraints would work., e.g., without group invariance.}
Analogously, we could write a reconstruction equation on the template if $\omega_A^g$ was invertible.

We emphasize that a psuedoinverse is not a satisfactory substitute for a true inverse. 
Our requirement, as shown earlier in Eqn. \eqref{eqn:recovery-rank-xi}, is $\omega_g^T$  be full rank.
Equivalently, for a given shape value $(r,\alpha)=\varphi^{s}(\theta)$, there must be a unique $\dot{g}$ such that Eqn. \eqref{eqn:crawler-block-cons} holds.
Ergo, we need to synthesize another constraint from an observation function $f:Q \to \Reals$, as in \S \ref{sec:missing-constraints}.
Recall, our perspective that we're able to do this is rooted in recovering a high codimension object -- namely, a trajectory on the template.
As we saw in \S \ref{sec:missing-constraints}, there is an open an dense set of  functions that are suitable.
In this case, we arbitrarily  choose
\begin{equation}\label{eqn:crawler-extended-constraint}
 f(x,y,\theta,r,\alpha)=x-\theta 
\end{equation}

As we will see in the sequel, our desired trajectory will obey $x,\theta$ symmetry, defining the constraint
\begin{equation}\label{eqn:crawler-extended-constraint}
\dot{x}-\dot{\theta} = 0
\end{equation}
We augment Eqn. \eqref{eqn:crawler-template-cons} with the row defined by Eqn. \eqref{eqn:crawler-extended-constraint} exactly as shown in Eqn. \eqref{eqn:augment-cons} of \S \ref{sec:missing-constraints} to get augmented form $\tilde{\omega}_T $ so that the sub-block $\tilde{\omega}_T^g \in GL(3,\Reals)$\footnote{We numerically verified that rank is three in a neighborhood of the goal trajectory}.
We thusly have 
\begin{align}
&\tilde{\omega}_T^g \cdot \dot{g} + \tilde{\omega}_T^{\left(r,\alpha\right)} \cdot \left(\dot{r},\dot{\alpha} \right) = 0 \label{eqn:crawler-full1} \\
&\omega_A^g \cdot \dot{g} + \omega_A^{\theta} \cdot \dot{\theta} = 0 \label{eqn:crawler-full2}
\end{align}

Our desired template reference trajectory is shown in Fig. \mref{fig:crawler-shape-ref} of the main text.
The resulting group velocity is shown in Fig. \mref{fig:crawler-tracking} under the ``desired'' label.
We denote this reference template signal as $\gamma(t) = (g_d(t), r_d(t), \alpha_d(t)),~t\in[0,1]$. 
An anchor trajectory that produces the desired phase reference is shown in Fig. \mref{fig:crawler-anchor-ref}.
The anchor angles used to generate this curve are denoted by $\theta_{org}(t) \in T^6,~t\in[0,1]$. 

%\begin{figure}[ht]
%	\centering
%	\begin{minipage}{0.45\textwidth}
%	\centering
%	\includegraphics[width=\linewidth]{crawler_shape_ref.png}
%	\caption{$r$ and $\alpha$ coordinates of the template along the $\theta_{org}$ curve, i.e. - the desired reference trajectory on the template.}
%	\label{fig:crawler-shape-ref}
%    \end{minipage}\hfill
%	\begin{minipage}{0.45\textwidth}
%	\centering
%	\includegraphics[width=\linewidth]{crawler-theta-org.png}
%	\caption{The six joint angles of $\theta_{org}$}
%	\label{fig:crawler-anchor-ref}
%	\end{minipage}
%\end{figure}
%
%\begin{figure}[ht]
%	\centering
%	\includegraphics[width=0.8\linewidth]{crawler_tracking.png}
%	\caption{The COM group velocity $\dot{g} = (v_x, v_y, v_{\theta})$ over time. The ``desired'' curve is the undamaged motion that we are aiming to recover. The blue ``old'' curve is the group velocity achieved if no recover strategy is attempted post damaged. The ``recovered`` trace is the performance after recovery.}
%	\label{fig:crawler-tracking}
%\end{figure}

We now assume that the $\theta_1$ actuator is jammed, so that $\forall t, \theta_1(t) = \theta_1(0)$. 
This is a constant for all gaits, as we have also assumed the initial condition is fixed.
If we did nothing, and simply played back the same $\theta_{org}$ with the $\theta_1$ acutator stuck, we obtain considerable error.
In Fig. \mref{fig:crawler-tracking} the curve labeled ``old'' illustrates the resulting group velocity of the COM should ``playback'' be attempted without recovery.
Ergo, attempting to improve this via exploiting the redundancy of the four-bar linkages is warrented.

As we have the model equations \eqref{eqn:crawler-full1} and \eqref{eqn:crawler-full2}, we can write an explicit system of differential equations to encode our recovery strategy.

We first substitute $\gamma$ in Eqn. \eqref{eqn:crawler-full1},
\begin{equation}\label{eqn:crawler-auto-eq}
\dot{g}_r(t) = \left(\tilde{\omega}_T^g\right)^{-1} \tilde{\omega}_T^{\left(r,\alpha\right)} \cdot(\dot{r_d}(t),\dot{\alpha_d}(t))
\end{equation}
This is an autonomous differential equation that we know our curve $\gamma$ satisfies.
Since we are seeking a $\theta(t) \neq \theta_{org}, t \in (0,1]$ that satisfies Eqn. \eqref{eqn:constraint-fit}, we substitute Eqn. \eqref{eqn:crawler-auto-eq} into Eqn. \eqref{eqn:crawler-full2} to get
\begin{equation}\label{eqn:crawler-anchor-auto-eq}
-\left(\omega_A^g \cdot \left(\tilde{\omega}_T^g\right)^{-1} \tilde{\omega}_T^{\left(r,\alpha\right)} \cdot(\dot{r_d}(t),\dot{\alpha_d}(t)) \right) =  \omega_A^{\theta} \cdot \dot{\theta}  
\end{equation} 
Observe that  Eqn. \eqref{eqn:crawler-anchor-auto-eq} is also an non-autonomous differential equation, as the coefficients $\omega_A^g$ and $\omega_A^{\theta}$ depend on $\theta$. 
We additionally require that Eqn. \eqref{eqn:phase-fit} is satisfied, so we write
\begin{equation*}
\varphi^s(\theta(t)) = (r_d(t), \alpha_d(t)) \implies D_{\theta(t)} \varphi^s \cdot \dot{\theta} (t) = \left(\dot{r_d}(t), \dot{\alpha_d}(t)\right)
\end{equation*}
We employ the Moore-Penrose psuedoinverse to write this as the non-autonomous differential equation
\begin{equation}\label{eqn:crawler-phase-proj}
\dot{\theta}(t) = D_{\theta(t)}^{\dagger} \varphi \cdot \left(\dot{r_d}(t), \dot{\alpha_d}(t)\right)
\end{equation}

Using the approach in Eqn. \eqref{eqn:crawler-phase-proj} and \eqref{eqn:crawler-anchor-auto-eq}, and the constraint that $e_1 \cdot \theta = 0$, we write a combined system as the following
\begin{equation}\label{eqn:crawler-nonauto}
\dot{\theta}=
\begin{bmatrix} \omega_A^{\theta} \\
D_{\theta(t)} \varphi \\
e_1
\end{bmatrix}^{\dagger}
\begin{bmatrix}
-\left(\omega_A^g \cdot \left(\tilde{\omega}_T^g\right)^{-1} \tilde{\omega}_T^{\left(r,\alpha\right)} \cdot(\dot{r_d}(t),\dot{\alpha_d}(t)) \right) \\
 \left(\dot{r_d}(t), \dot{\alpha_d}(t)\right)\\
 0
\end{bmatrix}
\end{equation}
Eqn. \eqref{eqn:crawler-nonauto} is an non-autonomous, differential equation. 
We implemented it numerically in Python 2.7.5 with the \verb|numpy| and 
\verb|scipy| numerical processing libraries.
Solutions to this equation exist given the full-rank assumption imposed on the constraint equations \citep[Thm. 2]{rakovcevic1997continuity}.

We cannot directly integrate Eqn. \eqref{eqn:crawler-nonauto}.
We require that the constraints are enforced - i.e., Eqns. \eqref{eqn:crawler-full1} and \eqref{eqn:crawler-full2} must hold \emph{exactly}.
Conventional integrators are approximating the true solution, so the error compounds as the integration time grows.
If the constraints are violated too badly, it's not reasonable to consider the resulting trajectory segment a meaningful solution.

We used the finite-step-size method \verb|lsoda| from the \verb|scipy.integrate.ode| package. 
Provided with $x_i, t_i$ and vector field $f$, it returns $x_{i+1}$, and is thus called in a loop to produce a sequence $\{t_i,x_i\}_{i=0}^{t_n}$. 
At each $x_i,t_i$, we projected $x_i$ onto the constraint manifold using Newton iteration to point $p_i$, then passed $p_i,x_i$. 
into $f$, i.e.
\begin{lstlisting}
    ti = 0
    xi = x0
    while t < t1:
       ti,xi =integrate(ti,xi,t+dt)
       ti,xi = projectConstraints(ti,xi)
\end{lstlisting}

Employing this crude form of geometric integration on Eqn. \eqref{eqn:crawler-nonauto}, we obtained a new $\theta_{rec}(t)$, $r(t)$, and $\alpha(t)$, shown in Figs. \mref{fig:crawler-shape-ref} and \mref{fig:crawler-anchor-new}.
Note that we cite Fig. \mref{fig:crawler-shape-ref} for the new phase curve as well.
This is inteitional, as the new and old phase curves are numerically identical. 
%\begin{figure}[ht]
%		\centering
%		\includegraphics[width=\linewidth]{crawler_theta_new.png}
%		\caption{}
%		\label{fig:crawler-anchor-new}
%\end{figure}

The performance of our recovered joint inputs is shown in Fig. \mref{fig:crawler-tracking} as the ``recovered'' trace. 
As we can see, it appears to recover the desired group velocity (and thus, group position) quite well, especially in relative performance to no recovery at all.

\subsection{Enepod Methods}\label{sec:enepod}

Our experimental setup consists of three major components: the robot itself, the motion tracking system, and the optimization software.

The robot, as described in the main text, is a collection of Robotis modules. 
It is not autonomous in any way -- each module tracks a reference signal provided from an external source.
The motion tracking system is a Qualisys network of Opus-series cameras.
In the capture arena, the system provides time-series position measurements of the retroreflective markers.
The Qualisys telemetry is collected by the optimization software package, which is responsible for both filtering the telemetry, updating the gait parameters, and driving the robot.

Our optimization problem occurs in two stages. 
The first stages measures our chosen observation functions on desired gait, and caches the functions to memeory.

In the second stage, we then re-initialized the robot with gait parameters that were \emph{not} a viable gait.
In this case, they left the robot completely stationary.
The goal of the optimization is to generate a new set of feasible gait parameters.
Unlike the proceeding cases, the parameter space is not restricted in any way. 
In this setting, we would expect that a successful optimization can \emph{exactly} recover the desired motion.

\subsubsection{Coordinates}\label{sec:enepod-coords}
Each marker has a unique name for which we will consistently reference.
The marker names are depicted in Fig. \mref{fig:Enepod-coords}.
A marker $A$  has $(A_x,A_y,A_z) \in \Reals^3$ spatial coordinates with respect to an arbitrary world frame that arises by calibrating the motion tracking system.

We treat the robot as a rigid body with the legs attached.
However, the spine is articulated, so we define a virtual rigid body $C$ 
using the markers S1, S4, VR, and VL. 
We denote the COM of C as $(C_x,C_y,C_z) \in \Reals^3$.
As we are using $C$ merely as a location, we assume each marker is unit mass, so that the COM is just the mean of the S1,S4,VR, and VL locations.

%The robot has coordinates for each leg, as shown in Fig. \ref{fig:enepod-coords}.
We will also numerically refer to an entire leg by an index, as shown in the figure. 
E.g, Leg $2$ refers to the entire assembly of the leg $ML$ and associated markers attached to it.
\subsubsection{Gait Space}
\begin{figure}[ht] 
	\centering
	\includegraphics[width=.75\linewidth]{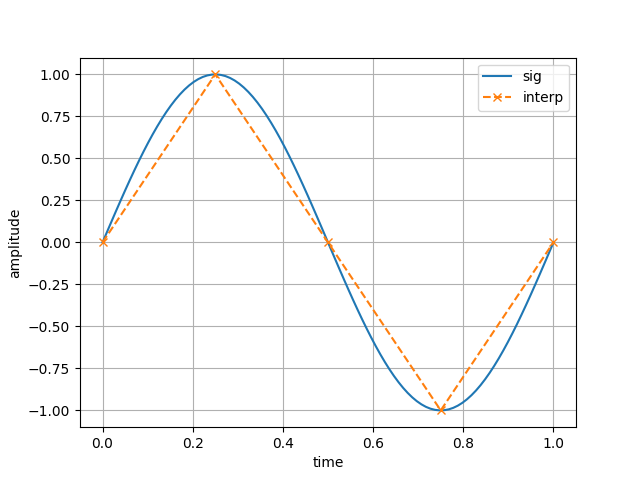} 
	\caption{Our parametric scheme for representing input signals. The 'sig' trace is a desired nonlinear function, while 'interp' is our linear interpolation of it. Our parameterization adjusts the amplitude of the knots, but not their horizontal location. The sawtooth approximation of a sine wave is operationally equivalent on the robot.}\label{fig:sawtooth-wave}
\end{figure}

There are seven Robotis modules, so we need to define seven input signals to fully realize a gait.
We represent this space with 28 parameters, four for each channel.
Each of the four parameters defines a knot point of a linearly interpolated signal. 
The $x$-coordinate of each knot point is fixed at (0, 0.25, 0.5, 0.75,1), respectively.
The last knot point at $x=1$ is set in software to equal the first, so that the resulting signal is periodic with $T=1$\footnote{The software rescales it to the desired frequency when driving the robot}.
The parameters then define the $y$-coordinates, allowing the representation of various ``triangle'' waves, as depicted in Fig. \ref{fig:sawtooth-wave}.
We chose this parameterization as our initial reference gait was composed of sinusoids. 
The mechanical bandwidth of the robot is such that the motion between re-sampling a sine wave with our parameter scheme, versus using a sine function is software, was indistinguishable. 
\subsubsection{Data Windowing}\label{sec:enepod-goal}
We are evaluate the cost function on the noisy data arising from the physical motion of the robot. 
We mitigate this by collecting enough data so that the mean of the cost function for a parameter set is distinguishable from the variance.
Ergo, we always take a window of $n=35$ strides (a stride is a single gait period), and average over them to get a mean value-per-stride (where ``value'' is the value of a function given via context)
As such, $t_0$ and $t_f$ are fixed at $t_0 = 0$, $t_f = 35 \times \frac{1}{2.5}=14$ seconds.

\subsubsection{Phase}\label{sec:enepodphase}

We operationally assume without proof that a phase-like quantity exists for the CT-SLIP, and produce a workable phase-map from hybrid data. 
As we merely require a submersion, rather than true conjugate dynamics, this should not be a major obstruction if trajectories are unique. 

The phase map $\varphi$ is estimated using the numerical tool \emph{Phaser} \citep{phaserRevzenGuck2008} to estimate a phase map. 
We operationally assumed that the tool will produce a workable phase map, ignoring any theoretical assumptions required (i.e. - we applied the tool to the measured data, and found the resulting function satisfactory).  
Recall, we only require \emph{any} submersion, rather than proper \emph{asymptotic phase}.

The utility \emph{Phaser} expects that the limit cycle of interest is contained within a two dimensional plane.
To satisfy this requirement, and to ensure that our phase map is a map from \emph{configuration} space, we project onto the first two principal components of the nominal gait position data, then ran phaser on the resulting trajectories.
The resulting phase map was cached for use in all evaluations of the cost function.

As this principal component projection occurs at every evaluation of the cost function, we chose an initial basis for this space when the nominal gait as executed. 
We then cached this choice of basis for subsequent projections, insisting that the subspace was well-defined and consistently oriented for all trials.

\subsubsection{Observation Functions}
There are four observation functions used - the vertical deflection of each tripod, and the horizontal deflection of each tripod.
We assume the legs have no meaningful mass in comparison to the spine, so that they contribute no momentum as masses in their own right. 
The legs only generate forces on the body through their deflection. 
We furthermore assume that the springs are perfectly one-dimensional - they are only able to deflect in their principal (softest) direction. 
Since stiffness scales exponentially with dimension, the legs are significantly stiffer on the axes off the major length.

We build a nominal value for the observation function by measuring its value while the robot is executing the goal behavior. 
By computing and caching Fourier-series approximations of the observation functions along the desired trajectory, we develop the constraints as we did in \ref{sec:missing-constraints}.
The quality-of-fit to the constraints for the disturbed robot will be applied as additional terms to the cost function shown in \S \ref{sec:enepod-goal}.
We will refer to the signals defined by the vertical and horizontal spring deflections along the nominal gait a $V_{ref}(\phi) \in \Reals^2, \phi \in [0,2\pi]$, and $H_{ref}(\phi)\in \Reals^2, \phi \in [0,2\pi])$, respectively.
When we evaluate these functions along the $i$-th iterate, we will use the notation $V_i$ and $H_i$, respectively.

The enepod is equipped with retroreflective markers that we use for motion capture experiments. 
We employ a Qualisys tracking system to measure position data.
Each marker produces a uniformly sampled telemetry stream of $(x,y,z)$ position data with respect to a fixed world frame.
The data is uniformly collected at 250 Hz.

\subsubsection{Vertical Spring Deflection}
The vertical spring deflection $V_i$ is calculated from telemetry by taking the arc cosine of the angle determined by the inner product $\langle SN-TN, nO-nI \rangle$, where $n$ denotes the leg index, as shown in Fig \mref{fig:Enepod-coords}.
I.e, 
\begin{align}\label{eqn:vert-1}
V_1 &:= \arccos \left(\langle S1-T1, FLO-FLI\rangle \right)\\ 
 &+\arccos\left(\langle S2-T2, MRO-MRI\rangle \right) \nonumber \\
&+\arccos \left(\langle S4-T4, BLO-FLI\rangle \right) \nonumber
\end{align}
\begin{align}\label{eqn:vert-2}
V_2 &:= \arccos \left(\langle S1-T1, FRO-FRI\rangle \right)\\  
&+\arccos\left(\langle S2-T2, MLO-MLI\rangle \right) \nonumber \\
&+\arccos \left(\langle S4-T4, BRO-FRI\rangle \right) \nonumber
\end{align}
We have assumed that the angle is sided (e.g, there are six distinct angles, as we are ignoring any geometric constraints across the body), despite the left and right angles resulting from the deflection of the same piece of spring steel as the crossbeam is clamped to the body in the middle.
Fig \ref{fig:vert-def} depicts an example time series collected from the enepod striding forward with its limbs cycling at 2 Hz.
While the x-axis is given in the number of sampled points rather than time, it was communicated with TCP and resampled, so that there are no missing packets and a constant $\Delta t$. 
The time series is the summed deflection for data of each tripod, i.e. - the three deflection signals were added together for each tripod, reducing the six signals to two - we expect this to be a reasonable aggregate, as the gait alternates which tripod is contacting the ground. 
It also suggests the redundancy we assumed the robot has initially; we are not interested in individual \emph{forces}, but rather we would like the total wrench induced by the legs on the spine to remain invariant.
Regardless of the number of legs employed, if the system remains controllable in the necessary directions, it would be possible to do so. 

Our computation procedure is as follows.
We first use the rigid body (as defined in \S \ref{sec:enepod-coords} to transform the marker data into the body frame.
In these coordinates, the body motion is periodic.
Our procedure is to slice an execution of the robot into strides, via the phase map computed in \S \ref{sec:enepodphase}. 
Computing the functions in Eqn. \eqref{eqn:vert-1}, and Eqn. \eqref{eqn:vert-2} on the nominal gait, we obtain the signals depicted in Fig. \mref{fig:vert-def}.
\begin{figure}[ht]
	\centering
	\begin{minipage}{0.45\textwidth}
		\centering
	     \includegraphics[width=\linewidth]{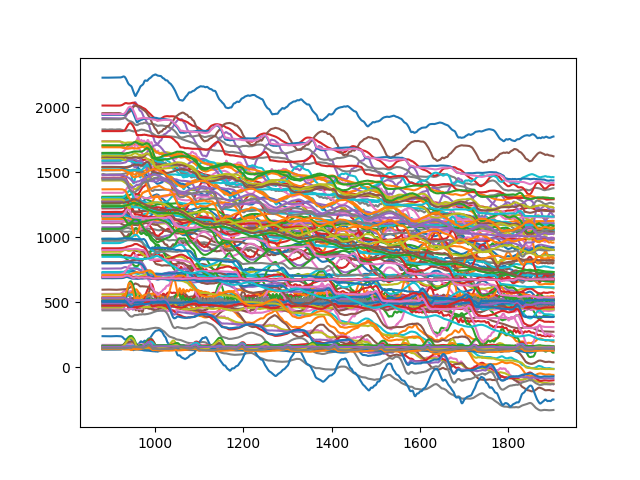} 
	    \caption{The filtered marker traces in world coordinates for an execution. The robot starts stationary, walks for a time, and then stops.}\label{fig:vert-def-mrks}
	\end{minipage}\hfill
	\begin{minipage}{0.45\textwidth}
		\centering
		\includegraphics[width=\linewidth]{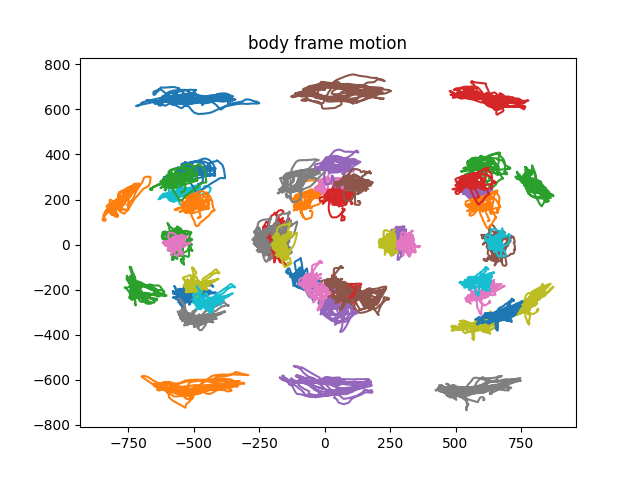}
		\caption{The same traces transformed into body coordinates using the virtual rigid mass.In these coordinates, the bais due to translation in the world frame is removed.}
		\label{fig:crawler-anchor-new}
	\end{minipage}
\end{figure}

\

%\begin{figure}[ht] 
%	\centering
%	\includegraphics[width=\linewidth]{figures/vrt_def_phase.png} 
%	\caption{The summed vertical deflection signals for tripod one and two. The dashes lines are over a single stride, while the solid lines are Fourier-series fits of the measured data.}\label{fig:vert-def-mrks}
%\end{figure}

\begin{figure}[ht] 
	\centering
	\includegraphics[width=.75\linewidth]{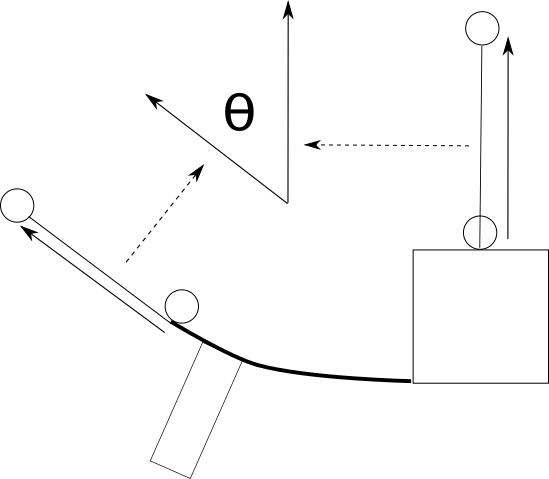} 
	\caption{A schematic of how the vertical deflection angle is determined from marker data. The distal markers are used as an angular gain to compensate for the resolution of the motion capture system.}\label{fig:vert-def-sch}
\end{figure}

There quite apparently seem to be two distinct periodic signals approximately $\pi$ out-of-phase with respect to each other, mod their magnitude.
The reason for the bias is magnitude from one tripod to another is unknown. 

\begin{figure}[ht] 
	\centering
	\includegraphics[width=.75\linewidth]{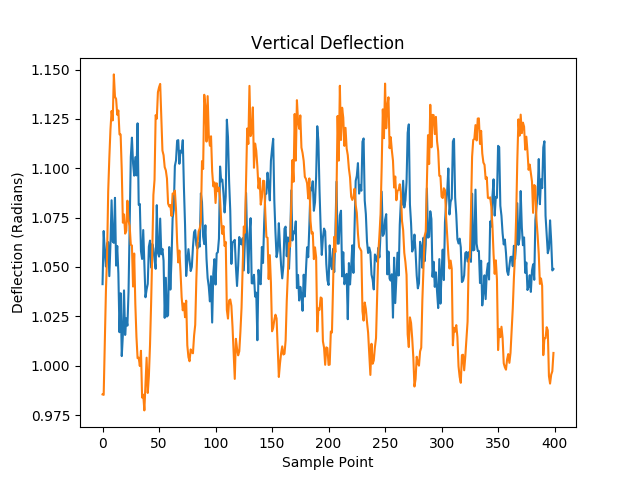} 
	\caption{Time series data -- the (BR,FR,ML) tripod signal is colored orange, while the (BL,MR,FL) tripod signal is shown in blue. The jump near data point 600 is where the robot changed directions. } \label{fig:vert-def}
\end{figure}

While the structure of the summed effect is clear, when re-visualized as a function of phase, an even more predictable signal emerges, shown in Fig. \mref{fig:vert-def}.
The tripod's vertical deflection seems to be functionally related to the phase estimate -i.e. - quite periodic.

\begin{figure}[ht] \label{fig:vrt-ph-def}
	\centering
	\includegraphics[width=\linewidth]{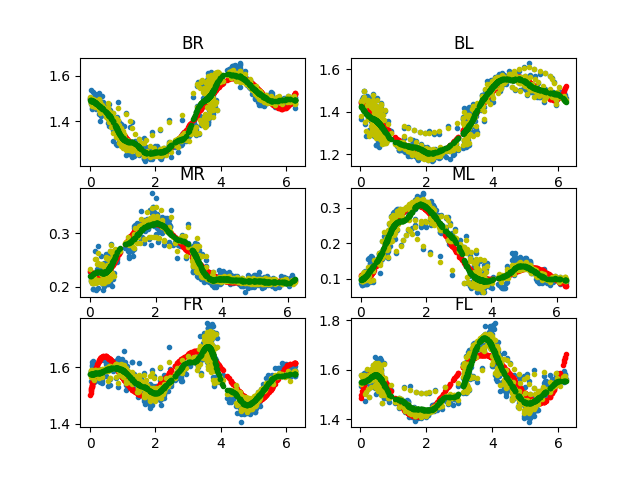} 
	\caption{Phase plots of the vertical spring deflection. The blue is the original data plotted as a function of phase. Yellow is after running a 2-order 2-sample discrete low pass filter twice, yellow is a 2nd order polynomial fit, while green is a 10th order Fourier series. We seem considerable symmetry between tripods. Note that the (BL, MR, FL) tripod has been phase-shifted by $pi$ for clarity. } 
\end{figure}
\subsubsection{Horizontal Spring Deflection}

The fore-aft deflection of the spring vertically-mounted springs (e.g, the deflection in the ``horizontal'' direction) was measured differently from the vertical deflection.
The marker sets indicated in Fig. \ref{fig:enepod} were used to define the two centroids - $C_{top}^i$, and $C_{foot}^i$, respectively, where $i$ indexes the leg.
We define the signal, where $\pi_{xy} : \Reals^3 \to \Reals^2$ is projection onto the $xy$-plane, by
\begin{equation}
fa^i(t) := \pi_{xy} \left(C_{top}^i -C_{foot}^i \right)
\end{equation}

\begin{figure}[ht]
	\centering
    \includegraphics[width=.75\linewidth]{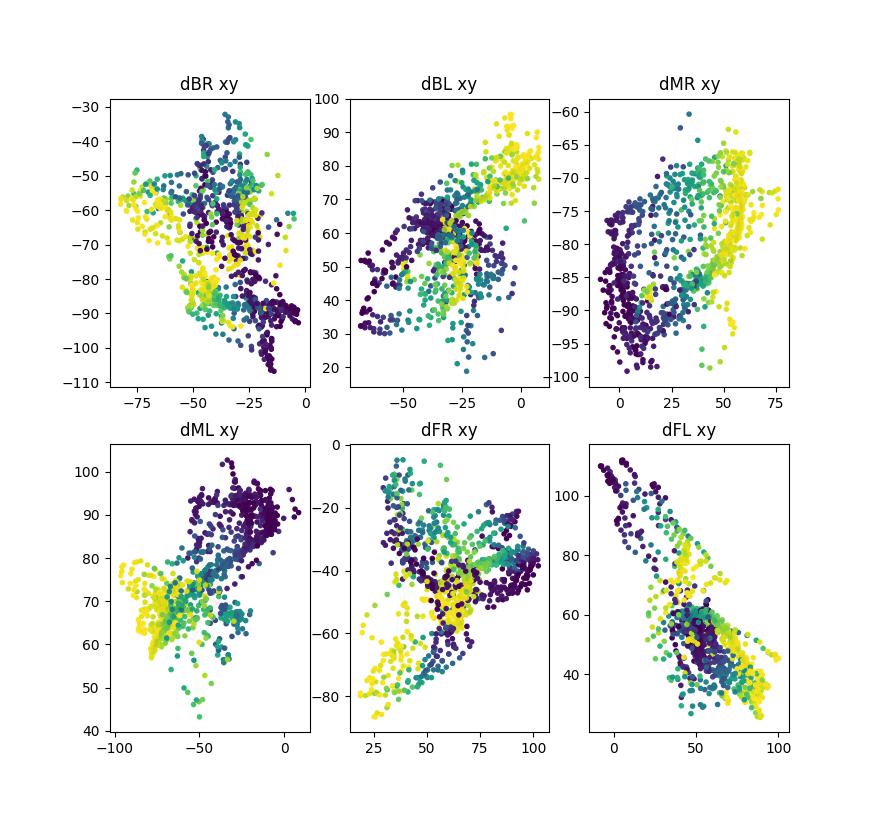}
    \caption{$fa^i(t)$ for each limb, organized by phase. Darker colors are earlier values of phase, while war colors are later phase values. We observe that there is not an apparent consistency between various limbs.}.\label{fig:enepod-hori-phase}
\end{figure}

\begin{figure}[ht]
	\centering
	\includegraphics[width=.75\linewidth]{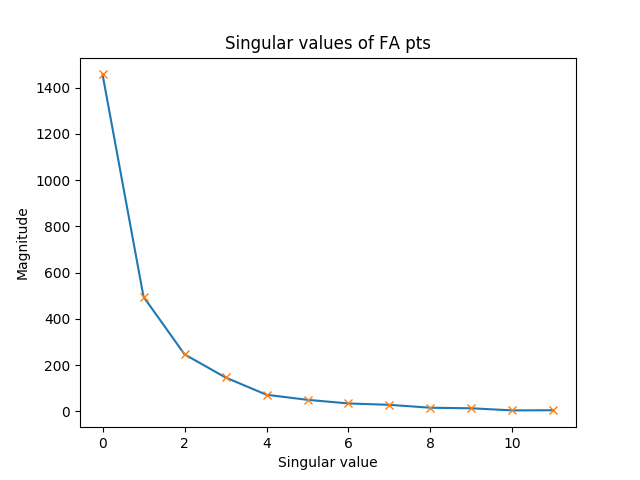}
	\caption{The singular values of the $fa^i$ data. It appears the data is predominately two-dimensional, but in such a way that does not correspond direct deflection of each tripod.}.\label{fig:enepod-hori-sing}
\end{figure}

If we merely repeated the analogous aggregating we did in Eqn. \eqref{eqn:vert-1} and \eqref{eqn:vert-2}, we did not see a meaningful signal. 
In Fig. \ref{fig:enepod-hori-phase}, we plot heatmaps of the $fa^i(t)$.
The $xy$ axes correspond to those coordinates.
Cooler colors are earlier phase values (closer to 0), while warmer colors are later phase values (closer to $2\pi))$.
We don't observe a meaningful correlation between the limbs.
As such, we instead aggregate data by projecting onto the first two principal values.
We argue that such a reduction is justified based on the singular value spectrum depicted in Fig. \ref{fig:enepod-hori-sing} -- the first two singular values summarize the data substantially. 
We plot the first two principal components of the nominal gait in Fig. \mref{fig:Enepod-hori-def}.

%\begin{figure}[ht]
%	\centering
%	\includegraphics[width=\linewidth]{fa_def_phase.png}
%	\caption{The first two principal components of the fore-aft deflection data. The dashed lines indicate measurement data, while the solid line is the Fourier-series approximation of the data.}\label{fig:enepod-hori-def}
%\end{figure}

\begin{figure}[ht]
	\centering
	\includegraphics[width=\linewidth]{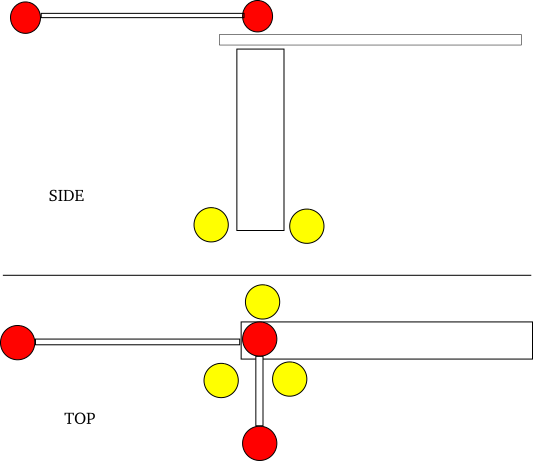}
	\caption{Sagittal and aerial view of one leg. The red and yellow groupings of markers are used to determine a relative deflection}
\end{figure}

\subsection{Asymptotic Phase for Limit Cycles}\label{appendix:phase-limit-cycle}
Since our primary interest is in steady gaits, which we assume are defined by limit cycles, Eqn. \eqref{eqn:def-phase} simplifies. 
Given an exponentially stable periodic solution, two points $x,y$ are considered to be ``asymptotically equivalent'' if and only if, for open set $D \subset \mathbb{R}$, vector field $f : D \to \T D$ with flow $\phi(x,t)$, and $x,y \in D$
\begin{equation}\label{eqn:def-cycle-phase}
\lim_{t \to \infty} \norm{ \phi(x,t) - \phi(y,t)} = 0 
\end{equation}

Denoting two such points via $x \sim y$, we observe that this is a equivalence relationship. 
As such, the equivalence classes partition $D$, and for each class, we denote a representative with $\theta (x) \in D$, and say $\theta(x)$ is the ``asymptotic phase'' or simply ``phase'' of $x$.

For $f \in \mathcal{C}^{\infty}$, and an exponential stable limit cycle for attractor, Winfree and Guckenheimer \citep{isochronsGucken1975} illustrated that these equivalence classes (which they dub as ``isochrons'' as they are permuted under the flow $\phi^t(x)$) have more structure. 
Denote the limit cycle of $f$ by $\Gamma \subset D$. 
No two points of $\Gamma$ are asymptotically equivalent, while every point in the stability basin $B$ of $\Gamma$ is equivalent to some point of $\Gamma$. 
Furthermore, each isochron is permuted under the flow, and are immersed submanifolds that foliate the stability basin of the attractor. The following theorem (\ref{thm:asym-phase}) , which is a restatement of Theorem 3 of \citet{ddfaRevzenKvalheim2015},
\begin{theorem}{\emph{Asymptotic Phase and Isochrons}}\label{thm:asym-phase} 
	
	$\forall \beta \in \Gamma$, let $S_{\beta} := \theta^{-1}(\beta)$. The isochron $S_{\beta}$ is a $C^{\infty}$ injective immersion of $\mathbb{R}^{n-1}$, and the map $\theta: B \to \Gamma$ is a $C^{\infty}$ submersion. Each isochron $S_{\beta}$ is transverse to $f$ , i.e, $\forall x \in S_{\beta}, f(x) \notin \T_x S_{\beta}$. Additionally, $B = \bigcup_{\beta \in \Gamma} S_{\beta}$, and $\phi(S_\theta(x), t) = S_{\theta( \phi(x,t))}$.
\end{theorem}

\emph{Remark} The map $\theta$ has an image that is diffeomorphic to $S^1$, which allows phase to be naturally represented with parameterizations of the circle, so that the ``phase of a point'' is not a point on $\Gamma$, but instead a real or complex number in the chosen coordinates of $S^1$.
In \citet{revzen2018temporal}, phase can be equivalently defined with respect to an arbitrary point on $\Gamma$, leading to the definition of $\tau : B \to [0,T)$, where $T$ is the period of $\Gamma$. 
While in this formulation, phase is a real number that denotes progress around the cycle from a reference point, it is not continuous at $0$. 
The issue of ``wrapping'' around the circle continuously can be resolved by considering $S^1 \subset \mathbb{C}$,  and defining $\phi : B \to S^1$, where $\phi(x) := \exp{ \left(i \omega \tau(x) \right)}$. 

$\theta(x)$ can be used to define the \emph{temporal 1 form} in the following fashion.
$\Gamma$ is one-dimensional.
Therefore, there exists a unique $\omega \in T^* \Gamma$ such that $\omega(f\left.\right|_{\Gamma}) = 1$. 
Then, we have the pullback $\theta^*\omega \in T^* \mathcal{B}$. 
By definition, $\theta^*\omega(f) = 1$ everywhere on $\mathcal{B}$. 
$\omega$ is a 1-form (The closed form $\omega$ is analogous gradient of asymptotic phase, but is not assumed to be exact) on $B$ that allows phase comparison of vector fields through sign and magnitude. $\omega(v) > 0$ would imply a curve is evolving in the same direction temporally as the isochrons; $\omega(v) < 0$ would show the opposite, for example.

%$\phi$ allows easy definition of $\omega : \mathcal{B} \to \mathbb{R},~ \omega(x) = \left(\varphi(x)\right)^{-1} \nabla \varphi (F(x,\mu, \lambda))$.

Geometrically, each isochron intersects the limit cycle once, the solutions starting on the same isochons asymptotically coalesce together as time flows forward. 
Given that the limit cycle is stable, we know points in the stability collapse onto $\Gamma$, but the phase map identifies a specific point on $\Gamma$ that points of the same phase evolve synchronously with, i.e, given a point $x(0) \in \ S_{\theta(x)} \setminus \Gamma$, $x(t)$ asymptotically collapses to and flows synchronously with the solution starting at $\theta(x) \in \Gamma$.

The foliation of the stability basin into forward-commuting sets highlights that, for a given oscillator, states have a temporal ordering that is preserved \emph{even off the limit cycle.} 
To have a specific periodic solution with desired dynamics, certain states come before other states, and through the perspective of recovery, suggests a recovery condition by preserving a sequence of motions.
One posture follows another, etc.

The above discussion of phase is historical, and restricted to smooth systems. 
The alert reader will note that the models used in legged locomotion are \emph{not} smooth - they are often hybrid systems with an aerial and ground domains, that are separated by a non-trivial reset map. 
While in each domain, the dynamics are smooth, the entire system is not.

%\GC{Shai -- how does this go these days?}
The extension of asymptotic phase to hybrid systems is an open question.
For classes of periodic hybrid system that have constant-rank Poincar\'e maps, Burden, Revzen, and Sastry \citep{burdenRevzenModel2015} show that the hybrid dynamics can be smoothed by forming the adjunction space with a smooth structure for which the hybrid dynamics have a smooth periodic push-forward. 
The smooth system may admit the existence of aysmptotic phase in the sense above, which could lifted back to the original hybrid system. 
However, the lift may not itself be smooth - the isochrons may have cusps induced by each hybrid transition. 
As isochrons are extended to the entire stability basin of an oscillator by flowing backwards through time \citep{isochronsGucken1975}, a potentially infinite number of non-smooth points in each isochron develop as it approaches the boundary of the basin. 
Such a difficulty might be resolved through restricting to a compact neighborhood of the basin, for which only a finite number of degenerate point may exist. 
In even a more general periodic hybrid systems, the limit definition in \ref{eqn:def-phase} may still be considered, but the nature of the associated equivalence classes (isochrons) is less obvious, if they exist. 
The point being, the specific nature of a isochron in a hybrid system is still unclear, though several potential avenues for generalization are present. 

\end{document}